%% file: iclr2022_conference.tex
\documentclass{article} 
\usepackage{iclr2022_conference,times}
\usepackage{hyperref}
\usepackage{url}
\usepackage{booktabs}
\usepackage{adjustbox}
\usepackage{graphicx}
\usepackage{algorithm}
\usepackage{algpseudocode}
\usepackage{caption}
\usepackage{subcaption}
\usepackage{wrapfig}
\usepackage{subcaption}
\usepackage{caption}
\graphicspath{{./images/}{./images/original_set}{./images/Mdetr_fail}{./images/Tbd_fail}{./images/State_fail}{./images/counter_examples}{./images/FiLM_fail}{./images/results}{./images/question_types}{./images/performance_plots}{./images/consistency_densities}{./images/drop_densities}{./images/drop_densities_pixels}{./images/additional}{./images/visfail}}

\usepackage{xspace}
\input{math_commands.tex}

\usepackage{enumitem}

\title{Measuring CLEVRness: Black-box Testing of Visual Reasoning Models}

\author{Spyridon Mouselinos \\
University of Warsaw \\
\texttt{s.mouselinos@uw.edu.pl} \\
\And
Henryk Michalewski \\
University of Warsaw, Google \\
\texttt{henrykm@google.com} \\
\AND
Mateusz Malinowski \\
DeepMind \\ 
\texttt{mateuszm@deepmind.com} \\
}

\iclrfinalcopy 
\begin{document}

\maketitle

\begin{abstract}
How can we measure the reasoning capabilities of intelligence systems? Visual question answering provides a convenient framework for testing the model's abilities by interrogating the model through questions about the scene. However, despite scores of various visual QA datasets and architectures, which sometimes yield even a super-human performance, the question of whether those architectures can actually reason remains open to debate.
To answer this, we extend the visual question answering framework and propose the following behavioral test in the form of a two-player game. We consider black-box neural models of CLEVR. These models are trained on a diagnostic dataset benchmarking reasoning. Next, we train an adversarial player that re-configures the scene to fool the CLEVR model. We show that CLEVR models, which otherwise could perform at a ``human level'', can easily be fooled by our agent. Our results 
put in doubt whether data-driven approaches can do reasoning without exploiting the numerous biases that are often present in those datasets. Finally, we also propose a controlled experiment measuring the efficiency of such models to learn and perform reasoning.
\end{abstract}

\section{Introduction}
Are our artificial intelligence systems capable of reasoning?  Or like \textit{Clever Hans}, they use various cues only tangentially related to the task and rely on rote memorization with poor generalization?~\citep{pfungst1911clever, Johnson_2017_CVPR} This work revisits such a question and proposes an interactive framework with the communication channel between two players. The first player, which reasoning capabilities we are about to test, performs  visual reasoning tasks, we call it \vqaagent. 
The second player, which we call the \adversarial, 
is manipulating the scene so that it \textit{fools} the first player even though those changes still lead to correct reasoning steps among humans. Both players interact with each other only through questions, answers and the visual scene as shown in Figure~\ref{two_agents}. 
If the \adversarial manipulating the scene causes the \vqaagent to change its answer even though the new scene is still valid for the same question and answer, it is then the reasoning failure. It is similar to the following situation. Imagine a box is placed between two spheres. If you ask a question, \textit{is there a box between two spheres?}, the answer should be positive. Now, if we move the box anywhere so it does not cross any of the spheres, and ask the same question, the response should remain unchanged. In other words, we postulate that reasoning outputs of agents need to be invariant under scene configurations that are consistent with the questions-answer pairs. Moreover, in the spirit of generic adversarial attacks, we seek configurations that also pose little if any reasoning challenges for humans.

We propose an automatic and agnostic pipeline to benchmark the reasoning capabilities of various models, only assuming they can communicate by answering questions about the scene. Due to the recent stream of research in vision-and-language \citep{VinVL,Defense,GraphReasoningVQA,VLBert,vcrcnn,kamath2021mdetr,lxmert,Uniter}, we believe there will be an increasing number of vision models that operate through language. Moreover, we also consider the visual question answering framework set-up as a two-player system as an excellent benchmarking pipeline. We perform all tests by scene manipulations and observing how a tested model behaves under such changes. The pipeline does not require any knowledge of the internals of the tested model. It also does not manipulate the sensory information of such a model, e.g., pixels in the images, and all the manipulations are physically meaningful.
Even though our current pipeline uses synthetic scenes as only those can easily be automatically manipulated, our results have also real-world ramifications. If models are susceptible to semantically \textit{meaningless} changes\footnote{Changes that are consistent with the task constraints.} in scene configurations, in a synthetic setting, there are valid concerns that real-world robots could also be prone to manipulation of objects in a room.
Finally, our work also questions the possibility of training and benchmarking networks in a purely data-driven and offline, static manner.

\paragraph{Contributions.}
The main contributions of our work could be summarized in three points. 

\textbf{\textit{First}}, we propose a strong \textit{black-box} adversarial test, which makes no assumptions about the underlying mechanics of a tested model, formulated as a game between two players. Our test does not require any direct access to the tested model, even through its sensory information. In particular, it does not require gradients, output probabilities, or any access to the perceived image. Our work also deviates from bounded perturbations and instead focuses on global scene manipulations that are still consistent with the task constraints, and can change the behavior of a tested model.\newline 
\textbf{\textit{Second}}, we reformulate visual reasoning by integrating visual question answering with zero-sum two-player game frameworks. Under our novel formulation, a visual and adversary agents compete against each other through content manipulation. We believe that this is an initial step towards more sophisticated frameworks that integrate computer vision with multi-agent systems.\newline
\textbf{\textit{Third}}, we explore the limits of the data-driven approaches in synthetic visual scenarios, and demonstrate that current CLEVR models are lacking the efficiency to learn robust reasoning steps.

\section{Related Work}
\label{sec:related_work}
Our work touches upon various research directions, which we briefly describe here.
\pp{Visual QA.} Introduced as a visual counter-part of the Turing Test~\citep{malinowski2014multi,geman2015visual}, it became a computer vision task that requires a holistic visual understanding. Many other works have extended the task to larger datasets or videos or adversarial train-test splits~\citep{Antol_2015_ICCV,agrawal2018don,tapaswi2016movieqa,yu2019activitynet}. More recently, we also observe the generalization of the task to become a part of the vision-plus-language suite of benchmarks~\citep{lu2019vilbert,chen2019uniter,wang2021simvlm}. \citet{Johnson_2017_CVPR} have introduced a synthetic variant of the Visual QA problem that is more focused on reasoning and the control of the experimentation. 
Although it was shown that traditional methods~\citep{Yang_2016_CVPR} are not enough to solve that dataset, newer methods can pass the human baseline on this task. These methods are trained to operate on pixels and text directly~\citep{Santoro_2018_NEURIPS, perez2018film, hudson2018compositional,malinowski2018learning,kamath2021mdetr} or they induce programs as an intermediate representation~\citep{johnson2017inferring, Mascharka_2018_CVPR,yi2018neural}. \citet{hudson2019gqa} have shown that Visual QA models lack some reasoning robustness but their approach is mostly linguistic and still static.  Some other works also observe issues with static evaluation and proposed more dynamic benchmarks, e.g., with a human-in-the-loop~\citep{khashabi2021genie,nieadversarial,li2021adversarial,sheng2021human}. 
We extend the line of research on Visual QA by reformulating it as an interactive two-agents system, and show that CLEVR under such a new setting is still an unsolved problem. Our work also differs from previous works in that it focuses on visual reasoning, is interactive and fully automatic.
\pp{Adversarial perturbations.}
\citet{szegedy2014intriguing,FGSM,kurakin2016adversarial_scale,CW} have introduced early methods that can `fool' networks by performing tiny, visually imperceptible, perturbations of the input image. 
\citet{moosavi2017universal} have also shown universal and untargeted adversarial perturbations that are more transferable than previous approaches. Most methods perform \textit{white-box} attacks, where an adversarial model has access to the target model's parameters or gradients. In \textit{black-box} attacks such access is removed and, e.g., only output probabilities are available~\citep{guo2019simple}. Other \textit{black-box} attacks involve the use of surrogates to estimate gradients of the real target~\citep{papernot2017practical,NEURIPS2019_32508f53} or gradient-free methods~\citep{GradientFree}. More related to our work, \citet{cheng2018query} describe a \textit{black-box} system, which only assumes access to outputs of the network. However, all the perturbations described above are continuous-based where individual pixels are changed independently. This leads to improbable images and assumes access to the sensory inputs of the neural network. In contrary, \textit{semantic} perturbations~\citep{joshi2019semantic,zeng2019adversarial}, operate on semantically meaningful chunks of the input, leading to more plausible and physically interpretable perturbations. However, they still assume direct access to either images or ideally differentiable renderers. We can interpret our work within the adversarial perturbations framework, where the adversarial model can change the original scene by its manipulations, with zero access to the target model, including its sensory information and not limited to tiny changes but instead to semantically and globally \textit{meaningless} modifications.
\pp{Reinforcement learning.} The game between the \vqaagent and \adversarial is a zero-sum two-player game. \citet{conneau2017word} use a conceptually similar two-player game to find a better word alignment between two languages without parallel corpus. Inspired by such a problem formulation, we consider our two-player game as a tool to achieve better reasoning models.
We optimize the \adversarial with Advantage Actor-Critic (A2C) algorithm~\citep{sutton,degris}. 
\pp{Probing and measuring intelligence.} Turing Test~\citep{turing2009computing} is considered to be among the earliest works on measuring the intelligence of artificial systems. Crucially to us, it is also formulated as a two-player game with an interrogation protocol. \citet{legg2008machine} systematizes and relates the notion of intelligence from different fields with the main focus on `universality'.
% Similarly, \citet{chollet2019measure} investigates various attempts at defining or measuring intelligence and highlights the role of skill-acquisition efficiency. 
\citet{Johnson_2017_CVPR} have transferred the notion of intelligence onto the visual ground and emphasized reasoning rather than universality. It seems there is no widely accepted dataset or definition that encompasses our intuitions about intelligence. However, there are a few recent directions showing the lack thereof, mainly highlighting bias amplification or absence of mathematical capabilities ~\citep{hendricks2018women, bhardwaj2021investigating, PiekosMM20}.

\section{Preliminaries}
\label{sec:preliminaries}
In this section, we explain briefly how the CLEVR dataset~\citep{Johnson_2017_CVPR} is constructed and introduce our notation and definitions.
\pp{CLEVR} is a synthetic visual question answering dataset introduced by~\citet{Johnson_2017_CVPR}, which consists of about 700k training and 150k validation image-question-answer triplets. Images are artificially constructed and rendered from scene graphs -- a special structure containing information about object attributes such as position or color. Such a scene graph is also used to synthesize the ground-truth question-answer pairs by expanding templates according to the depth-first-search ordering. Ambiguous scenes are rejected. Each image represents an isometric view of the scene containing from two to ten objects. There are three classes of objects, \textit{spheres}, \textit{cubes} and \textit{cylinders}. Each object can also be either large or small and has one color out of four (brown, purple, cyan, yellow). It can also be either metallic or rubber-made. Every object has $x$ and $y$ coordinates that are confined within the $(-3, +3)$ range. We use the same generation process to render modified scenes.
\pp{CLEVR models.} Various models have been introduced to work with the CLEVR dataset, some even `solving' the dataset by achieving near perfect performance. Despite the strong offline performance, we test if those models' performance perpetuates in the more interactive setting where
configurations of the scene could be changed. Whenever possible, we use pre-trained CLEVR models. Otherwise, we train the remaining models from scratch by making sure we achieve results similar to published accuracy numbers on the validation set. We summarize all the models in Table~\ref{model_table}. We show the accuracy on the CLEVR dataset (\textit{Accuracy}), indicate if an architecture is trained from scratch (\textit{Re-trained}), briefly describe how multi-modal fusion and reasoning is conducted (\textit{Reasoning Mechanism}), and indicate any extra privileged information required during the training process (\textit{Extra}). For instance, some models require extra access to functional programs used during the dataset generation, use scene graphs as a supervisory signal (states), or always operate on scene graphs (input-states). Otherwise, the models were trained only from image-question-answer triples.
\pp{\minigames.} 
We formulate our problem as a \game between two players, \vqaagent and \adversarial.
The \vqaagent takes as input question-image pairs and provide answers to such questions. Some models use states (scene-graphs) that replace images or require programs~\citep{Johnson_2017_CVPR}. The whole game consists of all CLEVR data points. For our purpose, we extend the notion of the \game into \minigames.
The rules of \minigames are identical to the whole \game. The only difference is that each \minigame operates on a subset of the CLEVR dataset. We define the size of a \minigame by the number of datapoints that are attached to that \minigame. We sample data points for each \minigame randomly and mutually exclusively.
\minigames have analogies in the adversarial perturbations literature. \minigames of size one resemble per-image adversarial perturbations~\citep{FGSM,Deepfool} whereas a \minigame that has all data points%% $that is the whole \game, i.e., $|\gamemath_j| = |\gamemath|$,
is similar to universal adversarial perturbations~\citep{moosavi2017universal}. In this work, we investigate various \minigame sizes but due to the sheer scale we were unable to use the whole game as the \minigame. Larger \minigames make the optimization process more difficult as the domain where the \adversarial needs to operate increases. The training is also much more time-consuming and a sequential process. Instead, we can train multiple players on different \minigames independently and thus massively. We leave the arduous training of the universal \adversarial on the whole \game as a possible future direction.

\begin{table}[!htbp]
\caption{CLEVR models that we use as \vqaagents.}
\label{model_table}
\begin{center}
\small
\scalebox{0.9}{
\begin{tabular}{lcccl}

Model Name &                Accuracy &  Re-trained &   Reasoning Mechanism & Extra Data\\
\midrule
SAN \citep{Yang_2016_CVPR}  &    72.1  &      &  Attention &   \\
FiLM \citep{perez2018film}  &  96.2    &  \checkmark    &   Feature Conditioning &  \\
RN \citep{Santoro_2018_NEURIPS}  &   93.2   &  \checkmark &  Relational &   \\
IEP \citep{johnson2017inferring}  &   96.9   &      &  Neural Program Induction & Programs \\
TbD \citep{Mascharka_2018_CVPR} &  99.1    &      &  Neural Program Induction  & Programs \\
Mdetr \citep{kamath2021mdetr} &  99.7    &      &   Multimodal Transformer Querying & States   \\
State-Input Transformer (ours) &  96.8    &   \checkmark   &   Cross Attention &   Input-States \\
\bottomrule
\end{tabular}
}
\end{center}
\end{table}

\section{Environment}
\label{sec:environment}
We need to ensure that \adversarials create valid scenes that are \textit{consistent} and \textit{in-distribution}. Both properties are guaranteed by our environment enforcers.
\pp{Consistency.} Since scene manipulation may change the answer for a given question, we need to ensure this does not happen. That is, the new scene is still \textit{consistent} with the question-answer pair. The question-relevance enforcer achieves that by running functional programs associated with each question~\citep{Johnson_2017_CVPR} on the modified scene-graph. Hence, it gets the new ground-truth answer. The enforcer rejects the new scene if that new answer differs from the previous one. In this way, it guarantees the newly generated scenes give the same answers as the original scenes on the same question. Thus, we can generate equivalent scenes containing the same objects that have identical answers for the same questions. Using that enforcer, we can test if the \vqaagent's answers are invariant under such an equivalent class of scenes.
\pp{In-distribution.}
Even with the question-relevance enforcer, the \adversarial may still produce undesired outputs. For instance, it can stretch the whole scene thus violating the scene boundaries from the original CLEVR dataset, making, e.g., everything to look very small (\Secref{sec:stretching_behavior} in the appendix). Although it is still an interesting form of adversarial scene manipulation, we focus rather on the \textit{in-distribution} scene manipulations that respect the original boundaries. To enforce that property, we use a scene-constraint enforcer that checks the boundaries of the scene.
Without that enforcer, the \adversarial would quickly resort to stretching the whole scenes, achieving a form of adversarial attack that uses distribution shifts rather than content manipulation. It does so, e.g., by moving the camera away until objects are barely visible. We give a few such examples in the appendix (\Secref{sec:stretching_behavior}).
\section{\adversarial}
\label{sec:adversarial}
\begin{figure*}[!b]
\begin{center}
\includegraphics[width=0.85\linewidth]{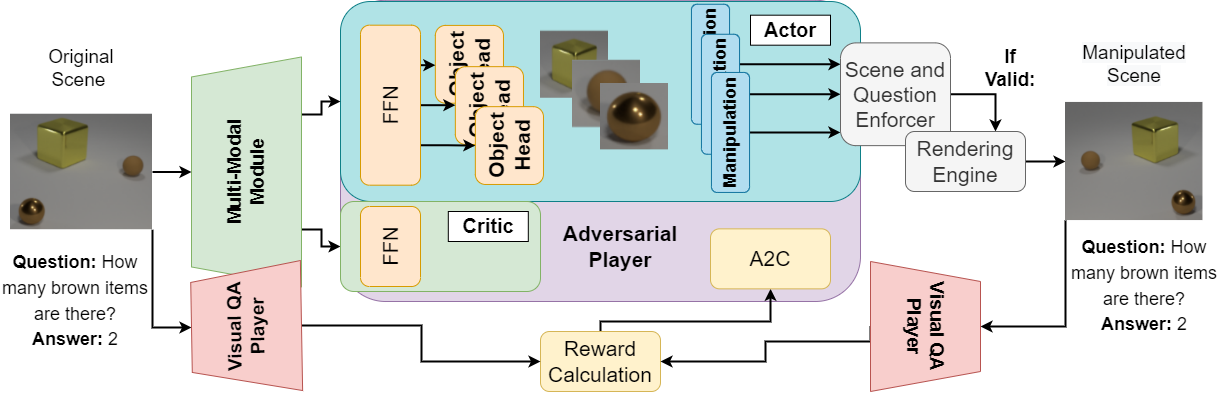} 
\end{center}
\caption{Our game between two players: \adversarial and \vqaagent. \adversarial uses a \textit{multi-modal module} to extract features conditioned on the visual and textual inputs. After transforming such features with a feed-forward architecture, it samples an action using object-specific heads. Each action corresponds to manipulating the corresponding object in the scene. In the case of missing objects, we use an $\emptyset$ token. After alternating the original scene graph, we use various environment enforcers to ensure validity of the constructed scene. A valid scene graph is rendered and introduced to the \vqaagent together with the original image. Finally, we collect responses of the \vqaagent and calculate suitable rewards based on them, and we repeat the whole cycle during the training phase.}
\label{two_agents}
\end{figure*}
Meaningful scene manipulations require not only generic scene understanding, but also the ability to distinguish which objects to displace and how. Hence, the player is a composition of a multi-modal module, which creates input representation, and a decision maker, which decides how to control the scene. 
Figure~\ref{two_agents} illustrates the \adversarial and the game between both players.
\pp{Multi-modal module.} We have experimented with the same multi-modal modules as those in \Tableref{model_table}, but found out we have a better performance and the convergence rate if the \adversarial operates on the scene-graphs (states) instead of pixels. For that, we use \textit{state-input} variant of Relation Networks~\citep{Santoro_2018_NEURIPS}. The model receives as the input $10 * 6$ object tokens, and question tokens. Every object token represents one-out-of-ten possible objects in the scene by its attributes such as position, color, shape, material and size. If the scene has fewer than ten objects, we use $\emptyset$ token to indicate that, which also acts as a padding. We also have special tokens that separate questions from the objects which we add as a latent embedding, e.g., $\text{emb}(\text{material}) + \text{emb}(\text{object})$. Such an input encoding is similar to our \textit{State-Input Transformer} and described in~\Secref{sec:state_input_transformer}. The embedded vectors are given to the Relation Network (RN). Finally, we train that network on the CLEVR visual question answering task, where we achieve $97.6\%$ on the validation set, and use the representation just after the last relational layer for the decision maker.
\pp{Decision maker.}
Inspired by the work on reinforcement learning \citep{a2c,conta2c,ddpg,sea2c,trpo}, we use an actor-critic module that acts on scenes. The actor is a general-purpose fully connected layer with ten object-specific heads. Each head is randomly assigned to a unique object in the scene for its manipulation. Every head produces a displacement in $x$ and $y$ coordinates of the corresponding object. Although we have initially experimented with the continuous output space, we have found out the following simple strategy is more effective. First, we discretize all the $x$ and $y$ coordinates into $N$ bins each. Now, each head produces two $N$-dimensional vectors that are next projected into a probabilistic space via softmax. Next, we sample displacements in $x$ and $y$ axis independently from both softmax distributions. Note that, even though we do not model the joint distribution explicitly due to computational reasons, both samples condition on the common head and thus are only \textit{conditionally} independent of each other. We discretize the scene where each axis has values in $[-3,3]$ onto $N=7$ bins per axis. Our critic is a simple three layer feed-forward network (with relu as activations) that predicts a reward score between $-1$ and $+1$ via tanh activation ($1.2*\text{tanh}$ for better numerical properties).
\pp{The game of scene manipulations.}
Due to our formulation of \adversarial and the environment, we can benchmark various reasoning models purely in the \textit{black-box} setting via a series of questions about the scene. \adversarial manipulates the scene so that it is still consistent with the question-answer pair. The manipulations are applied to scene-graphs, and the resulting scene-graph is evaluated by the environment enforcers described in~\Secref{sec:environment}. Invalid scenes are thus discarded. In this way, we ensure the \textit{in-distribution} and \textit{consistency} in the scene generation. Original image-question pairs are fed to a \vqaagent that produces corresponding answers. We refer to that answers as \oldanswers. After the scene manipulation, new images paired with the same questions are also given to the \vqaagent that produces \newanswers. We construct rewards based on \oldanswers, \newanswers and \textit{ground-truth answers}.

If the \adversarial forces the \vqaagent to change the answer, i.e., an \oldanswer is different than  a \newanswer, it gets \textit{Consistency Drop Reward} (\textit{cr}). If an \oldanswer is the \textit{ground-truth} answer, it gets instead \textit{Accuracy Drop Reward} (\textit{dr}). Both rewards differentiate between simply confusing the model and causing a drop in its performance. If \adversarial produces an invalid scene it gets \textit{Invalid Scene Reward} (\textit{isr}). This reward encourages producing scenes that pass the environment enforcers tests. Finally, if \adversarial does not manage to fool the model, it gets \textit{Fail Reward} (\textit{fr}). We use the following values: \textit{dr}$ = 1$, \textit{cr}$ = 0.1$, \textit{fr} $= -0.1$, \textit{isr} $= -0.8$.
\pp{Training algorithm.}
To train \adversarial we use the A2C algorithm %~\citep{Williams92simplestatistical} 
with the episode length set to one as we do not need to model long-range consequences of the decision-making mechanism. Batches contain images, question, answers, programs and scene-graphs. We train the \adversarial for each \minigame independently using the same architecture. We experiment with the following \minigame sizes $10, 100, 1000$. All \minigames are constructed randomly. Under our discretization scheme, the action space is $N^k$ where $N$ is the number of bins and $k$ is the number of objects in the scene. In practice, it is up to $49^{10}$.

\section{Experiments}
\label{sec:experiments}
\begin{figure}[!h]
\centering
\includegraphics[width=0.7\textwidth]{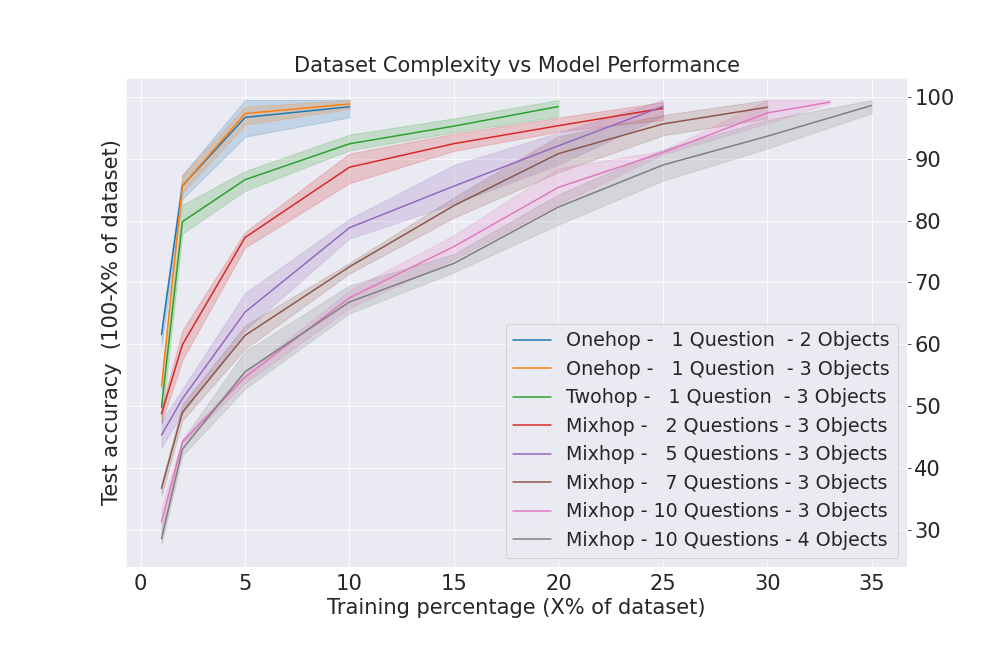} 
\caption{Data-driven limitation experiment: Unseen perturbation accuracy depends heavily on the number and complexity of introduced questions, followed by the number of objects in the dataset examples/scenes. A model could be considered robust when achieving close to 100\% test accuracy.}
\label{fig:generalization}
\end{figure}

\begin{figure*}[!htpb]
\centering
\small
\begin{adjustbox}{max width=0.99\textwidth, height=310pt}
\begin{tabular}{l@{\ }l@{\ }c@{\ }c@{\ }c@{\ }c}
 \toprule
 & \multicolumn{1}{c}{What is the color} &  
\multicolumn{1}{c}{How many other objects} & 
\multicolumn{1}{c}{How many other things are  } \\
& \multicolumn{1}{c}{of the block that is} & 
\multicolumn{1}{c}{are the same shape as the} & 
\multicolumn{1}{c}{the same shape as the big} \\
& \multicolumn{1}{c}{left of the big } & 
\multicolumn{1}{c}{big gray matte object?} & 
\multicolumn{1}{c}{metallic thing?}\\
& \multicolumn{1}{c}{blue matte ball?} & 
\multicolumn{1}{c}{} & 
\multicolumn{1}{c}{}
\\
\rotatebox{90}{FiLM - Before} & \multicolumn{1}{c}{\includegraphics[width=0.20\linewidth]{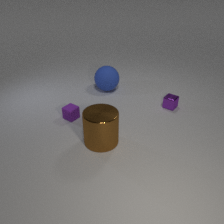}} &
\includegraphics[width=0.20\linewidth]{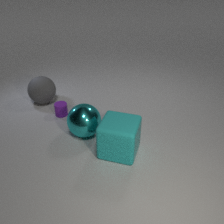} &
\includegraphics[width=0.20\linewidth]{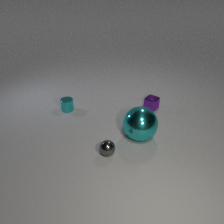}
\\
 & \multicolumn{1}{c}{\textcolor{green}{Purple}} & 
\multicolumn{1}{c}{\textcolor{green}{1}} & 
\multicolumn{1}{c}{\textcolor{green}{1}}
\\
\rotatebox{90}{FiLM - After} & \multicolumn{1}{c}{\includegraphics[width=0.20\linewidth]{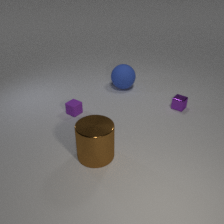}} &
\includegraphics[width=0.20\linewidth]{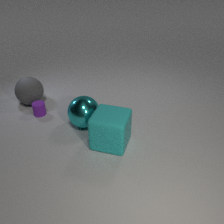} &
\includegraphics[width=0.20\linewidth]{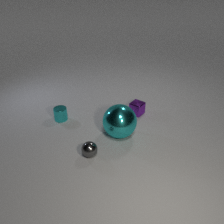}
\\
 & \multicolumn{1}{c}{\textcolor{red}{Brown}} & 
\multicolumn{1}{c}{\textcolor{red}{0}} & 
\multicolumn{1}{c}{\textcolor{red}{2}}
\\\midrule
 & \multicolumn{1}{c}{What number of tiny } & 
\multicolumn{1}{c}{Are there any other things  } & 
\multicolumn{1}{c}{Is there a big brown object  } \\
& \multicolumn{1}{c}{matte cylinders} & 
\multicolumn{1}{c}{that have the same material as} & 
\multicolumn{1}{c}{of the same shape as the} \\
& \multicolumn{1}{c}{are there?} & 
\multicolumn{1}{c}{the big yellow ball?} & 
\multicolumn{1}{c}{green thing?}
\\
\rotatebox{90}{TbD - Before} & \multicolumn{1}{c}{\includegraphics[width=0.20\linewidth]{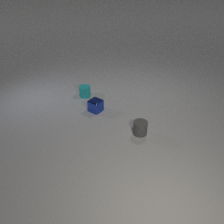}} &
\includegraphics[width=0.20\linewidth]{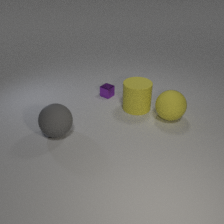} &
\includegraphics[width=0.20\linewidth]{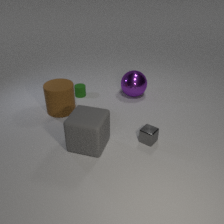}
\\
 & \multicolumn{1}{c}{\textcolor{green}{2}} & 
\multicolumn{1}{c}{\textcolor{green}{Yes}} & 
\multicolumn{1}{c}{\textcolor{green}{Yes}}
\\
\rotatebox{90}{TbD - After} & \multicolumn{1}{c}{\includegraphics[width=0.20\linewidth]{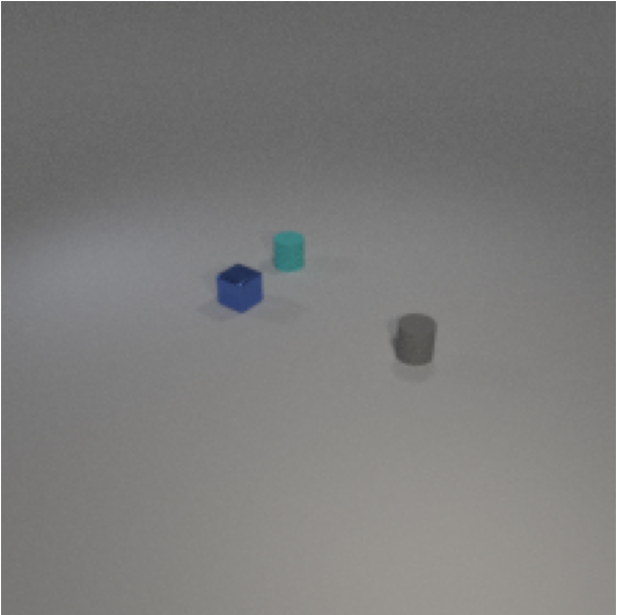}} &
\includegraphics[width=0.20\linewidth]{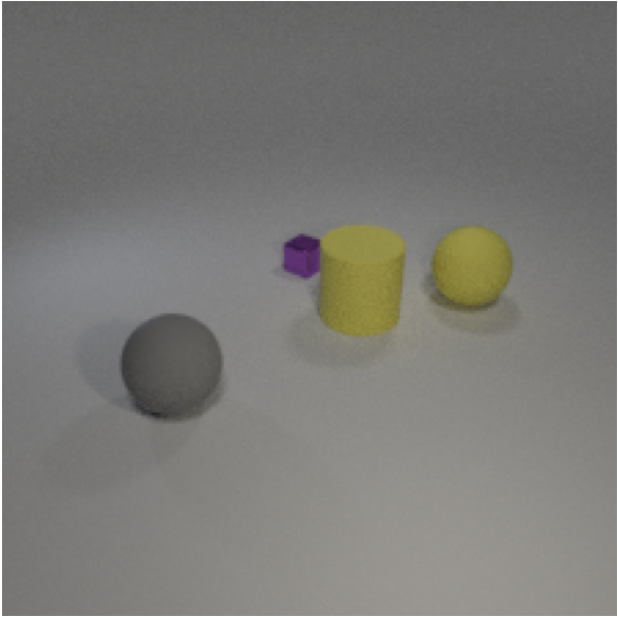} &
\includegraphics[width=0.20\linewidth]{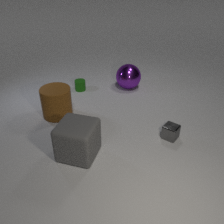}
\\
 & \multicolumn{1}{c}{\textcolor{red}{1}} & 
\multicolumn{1}{c}{\textcolor{red}{No}} & 
\multicolumn{1}{c}{\textcolor{red}{No}}
\\\midrule
& \multicolumn{1}{c}{The red thing that is } & 
\multicolumn{1}{c}{Is there anything else} & 
\multicolumn{1}{c}{\small What number of things are large} \\
& \multicolumn{1}{c}{the same material as the} & 
\multicolumn{1}{c}{that has the same material as} & 
\multicolumn{1}{c}{\small metallic blocks on the } \\
& \multicolumn{1}{c}{brown cube is what size?} & 
\multicolumn{1}{c}{the large green cube?} & 
\multicolumn{1}{c}{\small left side of the brown metallic } \\ 
& \multicolumn{1}{c}{} & 
\multicolumn{1}{c}{} & 
\multicolumn{1}{c}{\small thing or big cyan shiny objects?} 
\\
\rotatebox{90}{MDetr - Before} & \multicolumn{1}{c}{\includegraphics[width=0.20\linewidth]{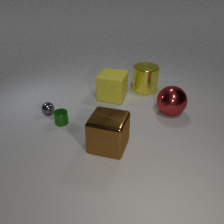}} &
\includegraphics[width=0.20\linewidth]{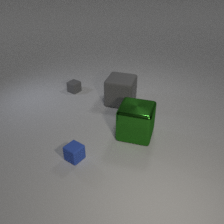} &
\includegraphics[width=0.20\linewidth]{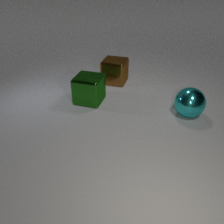}
\\
 & \multicolumn{1}{c}{\textcolor{green}{Large}} & 
\multicolumn{1}{c}{\textcolor{green}{No}} & 
\multicolumn{1}{c}{\textcolor{green}{2}}
\\
\rotatebox{90}{MDetr - After} & \multicolumn{1}{c}{\includegraphics[width=0.20\linewidth]{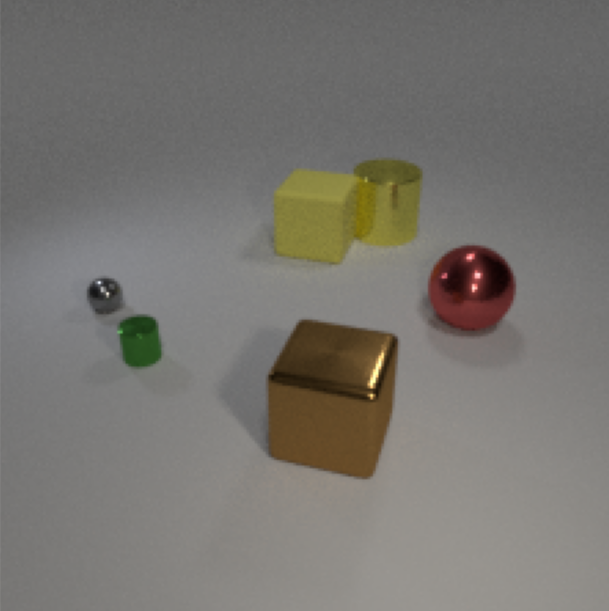}} &
\includegraphics[width=0.20\linewidth]{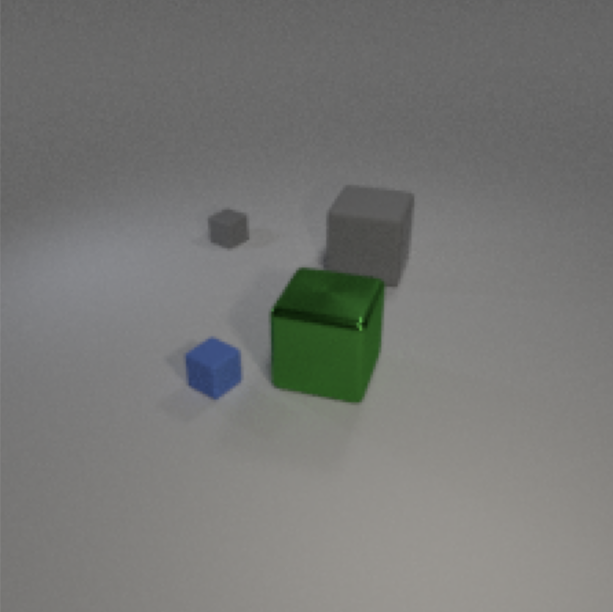} &
\includegraphics[width=0.20\linewidth]{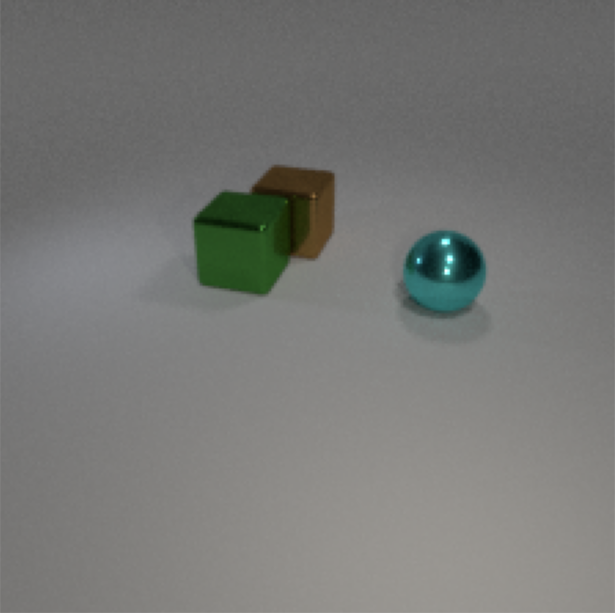}
\\
 & \multicolumn{1}{c}{\textcolor{red}{Small}} & 
\multicolumn{1}{c}{\textcolor{red}{Yes}} & 
\multicolumn{1}{c}{\textcolor{red}{0}}
\\\bottomrule
\end{tabular}
\end{adjustbox}

\caption{Manipulations of CLEVR models. We show results \textit{before} and \textit{after} scene manipulations.}
\label{fig:models_qualitative}
\end{figure*}

We consider \adversarial with a multi-modal module pre-trained either on states (\textit{state-input}) or directly from pixels (\textit{pixel-input}). We train \adversarial for each \minigame independently. We use different \minigame sizes in our experiments: $10$, $100$ and $1000$.
To obtain statistically significant results, we run each \minigame thirty times with randomly initialized \adversarials, each with different seed per trial. We report the results that are averaged over all those runs. We compute two metrics for each \minigame. \textit{Consistency} refers to the fraction of times \adversarial has changed the \vqaagent's answer, regardless if that answer was correct. \textit{Drop} refers to the fraction of times \adversarial has changed the \textit{correct} \vqaagent's answer. We aggregate the results using two statistics. \textit{Average Accuracy} averages accuracies over all \minigames. \textit{Maximal Accuracy} refers to the worst (best) case of a single \minigame from the \vqaagent's (\adversarial's) perspective.
We also computed $p$-values of T-Test, with the null hypothesis that a manipulation is unsuccessful. In almost all cases we reject the null hypothesis with small $p$-values.
The $p$-values are available in the appendix (\Secref{sec:quantitative_p_value}).
\pp{Quantitative results.}
We show our results in Table~\ref{quantitative_results} for the \textit{state-input} and \textit{pixel-input} \adversarials, and for each \vqaagent presented in \Tableref{model_table}. All the \vqaagents in the comparison are trained from pixels, apart form our custom \textit{State-Input Transformer} architecture (\Secref{sec:state_input_transformer} in the appendix). When comparing both \adversarials, we can observe that the \textit{state-input} \adversarial is significantly more successful than the \textit{pixel-input} one.  We hypothesize that this is the effect of having more structured and less ambiguous input information. It could also be the case that multi-modal modules trained from pixels ignore features in the same way and hence smaller discrepancy between all models operating on pixels. 
Moreover, as the size of \minigames increases, it is becoming more difficult for the \adversarial to successfully manipulate the model on all examples from the \minigame. This is an expected behavior resembling standard adversarial perturbations~\citep{FGSM} and their universal variants~\citep{moosavi2017universal}. The highest performing CLEVR models, TbD and MDetr, are the most robust under the manipulation. However, they are still susceptible to the scene manipulations in some \minigames as the \textit{Maximal Consistency} metric indicates, especially if our \adversarial uses states as the input. \Secref{sec:performance_distribution} (appendix) also shows richer performance statistics. Our detailed investigation shows that the models above are particularly sensitive to manipulations of scenes associated with \textit{counting} or \textit{existence} questions. Such question are often more complex, hence, increasing the likelihood of the reasoning failure at some stage (\Secref{sec:sensitivity_question_types} in the appendix).
\begin{table*}[t]
\begin{center}
\caption{Results of the games: \textit{Model dataset performance} refers to models' accuracy on the CLEVR dataset. \textit{Model \minigame performance} refers to the model accuracy on the \minigame examples where the largest (maximal) performance drop was detected, before manipulations took effect. The \textit{Average Accuracy} column reports the performance of models averaged over all runs of the respective \minigame sizes. \textit{Maximal Accuracy} reports the worst model performance among the respective \minigame size runs.
We also report in brackets the relative performance drop, in percentages $\frac{X-Y}{X}\%$. Average Accuracy is compared against \textit{Model dataset performance} ($X=$ Model dataset performance / $Y=$ State/Pixel-Input). Maximal Accuracy is compared against \textit{Model \minigame performance} ($X=$ Model \minigame performance, $Y=$ State/Pixel-Input).
Note that in the case of state-input and pixel-input \adversarial the worst \minigame might not come from the same \minigame instance.
}
\label{state_minigame_table}
\small
\resizebox{0.98\columnwidth}{!}{%
\begin{tabular}{p{0.8cm} p{1cm} p{1.5cm} p{2cm} p{2cm} p{2cm} p{2cm} p{2cm} p {2cm}}
\toprule
\multicolumn{1}{c}{} & \multicolumn{2}{c}{} & \multicolumn{2}{c}{Average Accuracy} & \multicolumn{2}{c}{Maximal Accuracy}\\
\vspace{5pt}Model &  \minigame size &\vspace{5pt} Model \newline dataset \newline performance  & \vspace{5pt} State-Input   &\vspace{5pt} Pixel-Input &\vspace{5pt} Model \newline \minigame \newline performance &\vspace{5pt} State-Input &\vspace{5pt} Pixel-Input\\
 &  &                  &           &     &                  &    &     \\
\midrule
SAN  & 10       &            &  61.8  (-14.2\%) & 63.8 (-11.4\%)     &  80.0 / 80.0      & 28.0 (-65.0\%)  & 43.0 (-46.2\%)  \\  
    & 100       &   72.1     &  66.1     (-8.3\%) &  69.2   (-3.9\%) &  74.0 / 75.0       &  47.6  (-35.6\%)   &   66.3 (-11.5\%)         \\
    & 1000      &            &  70.2       (-2.5\%) &  71.0  (-1.4\%) &  72.3 / 72.3      & 68.4  (-5.3\%)   &   70.8  (-1.9\%)        \\\cmidrule{1-8}

FiLM  & 10      &               &   83.9 (-12.7\%) &  93.6   (-2.6\%) & 100.0 / 100.0        &  48.0  (-52.0\%)   &    86.0        (-14.0\%)      \\
    & 100       &    96.2      &   89.1   (-7.3\%) & 94.8   (-1.4\%) &  98.0 / 100.0        &  75.6 (-22.8\%)   &    92.4    (-7.5\%)        \\
    & 1000      &              &    93.8  (-2.4\%) &  95.8  (-0.4\%) &  96.4 / 96.1      &   90.8  (-5.7\%)    &   94.7    (-1.4\%)      \\\cmidrule{1-8}

RN  & 10       &         &   80.5  (-13.6\%) &  86.4  (-7.2\%)        & 100.0 / 100.0         &   47.0 (-53.0\%)    &  74.0  (-26.0\%)               \\
    & 100      &    93.2     &  85.5  (-8.2\%) &  90.8  (-2.5\%)     &  94.0 / 95.0        &     63.2 (-32.6\%)   &  87.4 (-7.9\%)               \\
    & 1000     &         &  90.5  (-2.8\%) &   91.6  (-1.7\%)        &  93.1 / 93.3        &   90.0  (-3.2\%)   &   91.0 (-2.4\%)         \\\cmidrule{1-8}
IEP & 10       &          &   84.3 (-13.0\%) &  94.4  (-2.5\%)      &  100.0 / 100.0       &    48.0 (-52.0\%) &  87.0   (-13.0\%)                \\
    & 100      &   96.9       &  90.3  (-6.8\%) &  95.6  (-1.3\%)  &    98.0 / 96.0       &   74.4 (-24.0\%) &   93.5   (-2.6\%)           \\
    & 1000     &          &  94.1  (-2.8\%) &  96.6   (-0.3\%)      &   97.1 / 97.3     &   93.9 (-3.2\%) &   95.7  (-1.6\%)    \\\cmidrule{1-8}

TbD & 10       &           &  94.0 (-5.1\%) &  99.1   (-0.0\%)      &  100.0 / 100.0        & 69.0  (-31.0\%) &    99.0     (-1.0\%)        \\
    & 100      &     99.1   &     96.6   (-2.5\%) &  98.7  (-0.4\%) &  100.0 / 100.0         &  91.1   (-8.8\%) &   98.0  (-2.0\%)         \\
    & 1000     &        &   98.0    (-1.1\%) &   99.0  (-0.1\%)      &  99.4 / 99.7      &   95.5 (-3.9\%) &   98.8  (-0.9\%)       \\\cmidrule{1-8}

Mdetr  & 10     &        &  93.7    (-6.0\%) &  99.7   (-0.0\%)    &  100.0 / 100.0          &   60.0 (-40.0\%) &    99.0   (-1.0\%)         \\
    & 100       &   99.7      &  96.1   (-3.6\%) &   98.6 (-1.1\%) &   100.0 / 100.0           &  86.4 (-13.5\%) &     97.8    (-2.2\%)         \\
    & 1000      &         &  98.5   (-1.2\%) &   99.4 (-0.3\%)     &   99.5 / 100.0         &     94.5 (-5.0\%)   &   99.1 (-0.9\%)       \\\cmidrule{1-8}
    
State & 10        &     &   89.3    (-7.7\%) &  96.0  (-0.8\%)     &   100.0 / 100.0           &   77.0 (-23.0\%) &  91.0   (-9.0\%)     \\
Input  & 100     &   96.8    &   94.7    (-2.1\%) &  96.1  (-0.7\%) &   97.0 / 99.0           &  92.6 (-4.5\%) &   95.0  (-3.9\%)               \\
Transf. & 1000    &      &   95.7    (-1.1\%) &  96.5 (-0.3\%)      &   97.2 / 96.4        &   95.1 (-2.1\%) & 95.8   (-0.5\%)              \\
\bottomrule
\end{tabular}
}
\label{quantitative_results}
\end{center}
\end{table*}
\pp{Qualitative results.}
Figure~\ref{fig:models_qualitative} shows qualitative results of the scene manipulations for the three highest performing and distinctive models from Table~\ref{model_table}; e.g., simple feed-forward or program induction models. For the reader's convenience, we provide in \Secref{sec:object_categories} (appendix) examples of CLEVR object attributes. We can observe that the scene manipulations are surprisingly minimal and are semantically \textit{meaningless}, i.e., they should not `fool' human players.
\pp{Reasoning steps and visual cues.}
To better understand the source of errors (reasoning steps vs visual cues), we have conducted the following experiments. In the first experiment, shown in \Secref{sec:failure} (appendix), we visualize the attention mechanisms of \vqaagents before and after the scene manipulations. We observe changes in the attention map of all models but MDetr. The results suggest, most models's perception is impacted by the manipulations. However, all objects are still correctly identified by MDetr, thus suggesting some issues are also stemming from the reasoning steps. 
In the second experiment, we deployed Slot-attention object detector~\citep{slot}, and show that the detector identifies the same properties before and after manipulations. This result (\Secref{sec:obj_detect}; appendix) shows that \adversarial does not produce `corner cases' of the perceptual system. We also have similar conclusions based on our small-scale human experiment (\Secref{sec:human}; appendix).
\pp{Adversarial training.}
We have adapted a widely used adversarial training schema~\citep{goodfellow2018making}, and included manipulated scenes in the training protocol of a \vqaagent. We use \textit{pixel-input} RN~\citep{Santoro_2018_NEURIPS} as the \vqaagent. Using this method, we have obtained only marginal improvements in robustness at the cost of a slight performance degradation on the original CLEVR dataset. 
\pp{Limitations of data-driven reasoning models.} 
Due to the data-driven nature of our \vqaagents, we pose the following question, How many examples of \adversarial manipulations would be enough for a visual reasoning model to train on, in order for it to be robust against any unseen ones?
Intuitively, if the network has enough capacity, and was trained on all possible data points (or its manipulations), it could rely on a look-up strategy to solve the problem accurately. We propose the following experiment to address such a question. We created a series of datasets as follows.  We treat a scene as a discrete $7\times7$ grid (manipulations are restricted only to this grid). Thus each dataset contains $49^2 = 2401$, $49^3 = 117 649$, $49^4=5 764 801$ scene combinations, covering all possible scene manipulations with two, three, and four objects respectively. In the case of four objects, due to computational reasons, we keep a single object stationary (thus it also has $49^3$ scene manipulations).
The set of questions associated with each image dataset requires the model to perform either one reasoning step (Onehop), two reasoning steps (Twohop) or are a mixture of both (Mixhop).
We use RN, FiLM and TbD as \vqaagents. We train them on $X\%$ data and next evaluate on $(100 - X)\%$ remaining, unseen data. We have conducted ten trials by forming $X\%$ training data randomly. We notice that the behavior of all tested models is similar so we report a joint average. Results are presented in Figure \ref{fig:generalization}. As the number of objects and diversity of question increases, so does the amount of training examples needed in order to achieve robustness. 

\section{Conclusions}
We present a novel approach to isolate and benchmark the reasoning capabilities of visual models by formulating a game between two players. Both players are independent and can only communicate through a common interface, an environment and question-answer pairs. One player, named \adversarial, is trying to manipulate the environment so that the second player, named \vqaagent, can't answer the given question anymore. We show that popular models with solid results on CLEVR are susceptible to scene manipulations, and their performance may degrade significantly.
Moreover, we also conducted a controlled study on the generalization capabilities of such models. We did so by creating synthetic and simplified datasets of all possible scene manipulations. We show that the existing models are susceptible to reasoning gaps and require extra data, proportional to their task complexity in order to generalize to all possible configurations. One possible direction to increase efficiency could be the use of stronger inductive biases. Another to change the training paradigm into a more interactive one. However, the environment is synthetic, and it could be that learning from real-world and multi-modal data is somewhat more efficient. Finally, we believe that melding two-player games with the visual question answering framework is both natural and beneficial for testing the reasoning capabilities and potentially can also be generalized to other settings.

\subsubsection*{Acknowledgments}
All experiments were performed using the Entropy cluster funded by
NVIDIA, Intel, the Polish National Science Center grant
UMO-2017/26/E/ST6/00622 and ERC Starting Grant TOTAL. The work of
Spyridon Mouselinos and Henryk Michalewski was supported by the Polish
National Science Center grant UMO-2018/29/B/ST6/02959.

\clearpage
\bibliography{iclr2022_conference}
\bibliographystyle{iclr2022_conference}

\appendix
\include{appendix}

\end{document}

%% file: math_commands.tex
\usepackage{amsmath,amsfonts,bm}

\def\Figref#1{Figure~\ref{#1}}

\def\Tableref#1{Table~\ref{#1}}
\def\secref#1{section~\ref{#1}}
\def\Secref#1{Section~\ref{#1}}
\def\eqref#1{equation~\ref{#1}}
\def\1{\bm{1}}

\DeclareMathAlphabet{\mathsfit}{\encodingdefault}{\sfdefault}{m}{sl}
\SetMathAlphabet{\mathsfit}{bold}{\encodingdefault}{\sfdefault}{bx}{n}

\newcommand{\adversarial}{\textit{Adversarial Player}\xspace}
\newcommand{\adversarials}{\textit{Adversarial Players}\xspace}
\newcommand{\vqaagent}{\textit{Visual-QA Player}\xspace}
\newcommand{\vqaagents}{\textit{Visual-QA Players}\xspace}
\newcommand{\game}{\textit{Game}\xspace}
\newcommand{\minigame}{\textit{Mini-game}\xspace}
\newcommand{\minigames}{\textit{Mini-games}\xspace}
\newcommand{\oldanswer}{\textit{old answer}\xspace}
\newcommand{\newanswer}{\textit{new answer}\xspace}
\newcommand{\oldanswers}{\textit{old answers}\xspace}
\newcommand{\newanswers}{\textit{new answers}\xspace}

\newcommand{\pp}[1]{\newline\noindent\textbf{#1}}

%% file: appendix.tex
\newpage
\section{Appendix}
\subsection{URLs of CLEVR Models}
Table~\ref{table:model_links} shows the URLs to models used in our investigations (also Table~\ref{model_table} in the main paper). We also report if we re-trained a model from scratch (type \textit{Architecture}) or used already trained models (type \textit{Model}). Please note that the latter type proves that our testing procedure is fully black-box.
\begin{table}[!ht]
\begin{center}
\small
\scalebox{0.9}{
\begin{tabular}{lcccl}

Model Name &  Link  &  Type \\
\midrule
SAN \citep{Yang_2016_CVPR}  &      https://github.com/facebookresearch/clevr-iep & Model \\
IEP \citep{johnson2017inferring}  &   https://github.com/facebookresearch/clevr-iep & Model\\
FiLM \citep{perez2018film}  &   https://github.com/ethanjperez/film & Architecture \\
RN \citep{Santoro_2018_NEURIPS}  &      https://github.com/mesnico/RelationNetworks-CLEVR & Architecture\\
TbD \citep{Mascharka_2018_CVPR} &   https://github.com/davidmascharka/tbd-nets & Model\\
Mdetr \citep{kamath2021mdetr} &   https://github.com/ashkamath/mdetr & Model\\
\bottomrule
\end{tabular}
}
\end{center}
\caption{URLs to Models. \textit{Architecture} denotes that we use the code but re-trained the model on CLEVR. \textit{Model} refers to already trained models.}
\label{table:model_links}
\end{table}

\subsection{Object Categories}
\label{sec:object_categories}
\begin{figure}[!ht]
\centering
\includegraphics[width=0.6\textwidth]{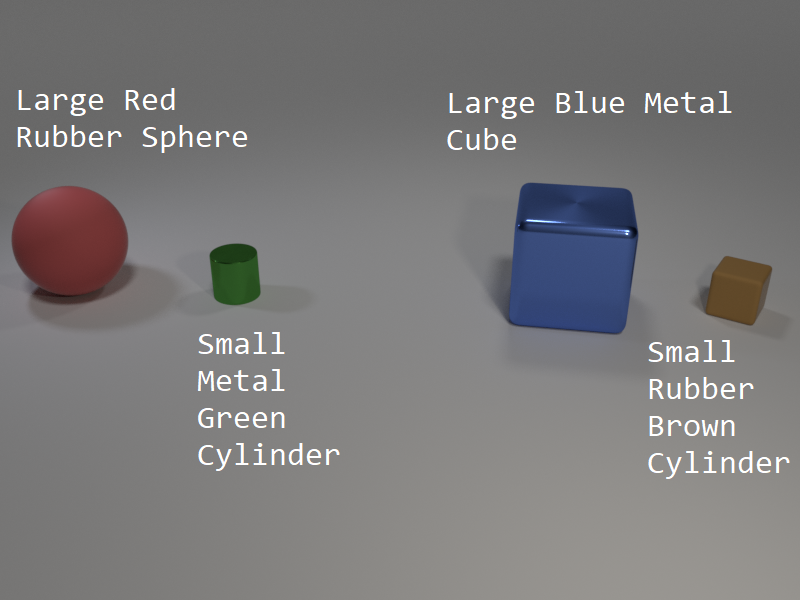} 
\caption{Object categories in the CLEVR dataset.}
\label{fig:prototypes}
\end{figure}

Figure~\ref{fig:prototypes} shows a made-up image that contains all available CLEVR object categories: shapes, sizes, and materials.
Small objects are rendered at approximately $40\%$ of the size of their large counterparts. Metallic materials can be identified by the high albedo property and `shininess' of the object. We provide it as the reference point for a reader to make it easier to compare our qualitative results.

\clearpage
\subsection{State-Input Transformer}
\label{sec:state_input_transformer}
\begin{figure}[!ht]
\centering
\includegraphics[width=0.8\textwidth]{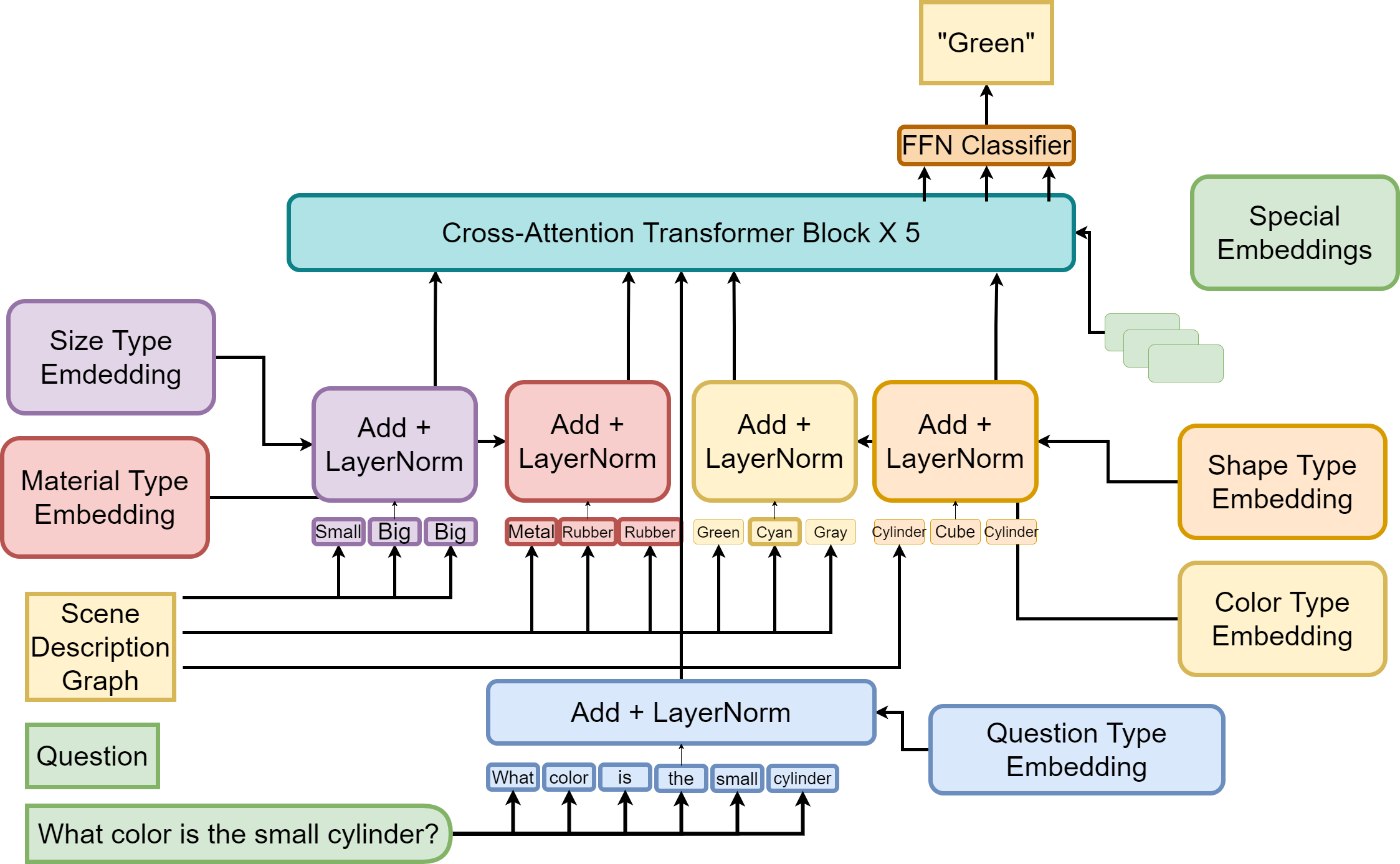} 
\caption{Our State-Input Transformer that operates on the graph scene (states).}
\label{fig:state_transformer}
\end{figure}

To test our two-player game against a \textit{state-input} \vqaagent, we have designed a transformer-based architecture~\citep{vaswani2017attention} that receives six types of input features -- object sizes, object shapes, object materials, object colors, object positions, and question tokens -- and uses a cross-modal attention mechanism. In that mechanism, queries from one modality attend to keys and values from the other modality.
Each data point contains variable-length inputs, describing all objects in the scene. It also contains question tokens that compose the question itself. Hence, we do padding with a special token $\emptyset$ to the maximal input length. We set the maximal length for the objects tokens to be $10 * 6 = 60$ and $50$ for the question tokens. All in all, we have $110$ input tokens.
Each token type is projected into a different embedding space and a learnable  type embedding is added to it, separately for object and question tokens. For instance, $\text{emb}(\text{material}) + \text{emb}(\text{object})$.  Furthermore, three special learnable embeddings are used as an additional input. We use them as queries to reduce the overall computational costs of the cross-attention mechanism. The same mechanism is also used in Perceiver~\citep{jaegle2021perceiver}. The input tokens (concatenated object and question tokens) form keys and values. In every transformer block, we apply cross-attention between all input embeddings and those three special tokens. This is repeated five times. As the final block, a feed-forward network (classifier) receives as inputs the three latent tokens and outputs answer probabilities. We show that architecture in Figure~\ref{fig:state_transformer}.

\subsection{Manipulations of \textit{State-Input} \vqaagent}
\begin{figure*}[!htpb]
\begin{adjustbox}{max width=0.95\linewidth}
\begin{tabular}{ccccll}
\toprule
& \multicolumn{1}{c}{\large Are there an equal number of small things that  \par} \\ & \multicolumn{1}{c}{\large  and large cyan matte balls? \par}\\
\rotatebox{90}{State - Before} & \includegraphics[width=0.3\linewidth]{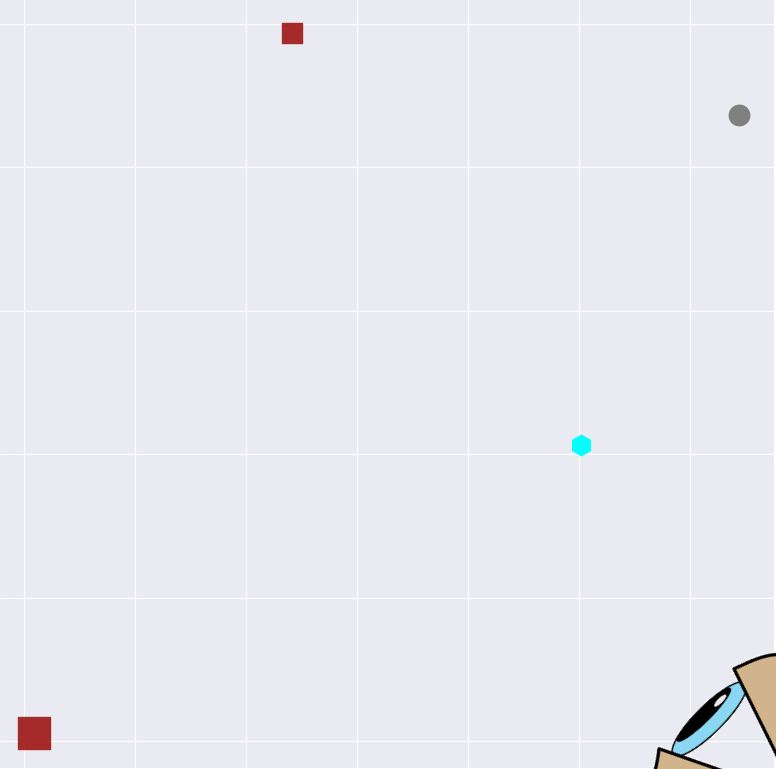} &
\includegraphics[width=0.3\linewidth]{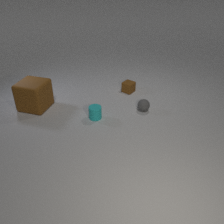} &
\\
 & \multicolumn{1}{c}{\textcolor{green}{}} & 
\multicolumn{1}{c}{\textcolor{green}{False}} & 
\\\midrule
\rotatebox{90}{State - After} & \includegraphics[width=0.3\linewidth]{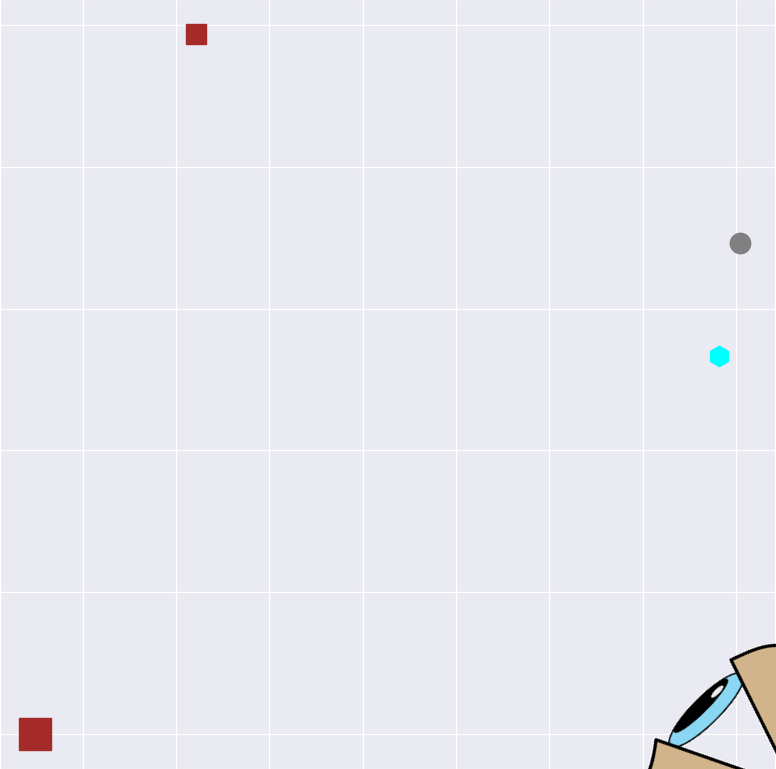} &
\includegraphics[width=0.3\linewidth]{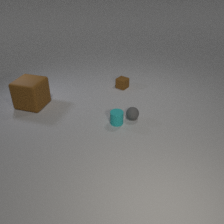} &
\\
 & \multicolumn{1}{c}{\textcolor{red}{}} & 
\multicolumn{1}{c}{\textcolor{red}{True}} & 
\\
& \multicolumn{1}{c}{\large Is the number of big blocks that are in front of the brown metallic cylinder \par} \\  & \multicolumn{1}{c}{\large the same as the number of tiny cylinders on the right side of the rubber cube? \par} \\

\rotatebox{90}{State - Before} & \multicolumn{1}{c}{\includegraphics[width=0.3\linewidth]{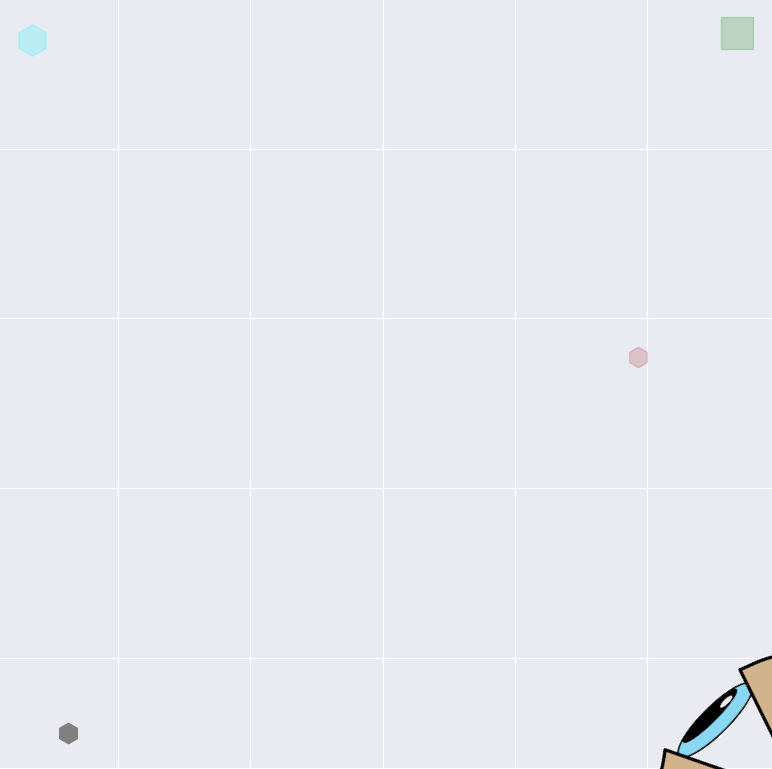}} &
\includegraphics[width=0.3\linewidth]{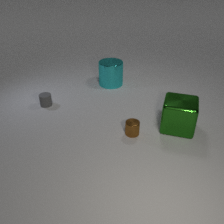} &
\\
 & \multicolumn{1}{c}{\textcolor{green}{}} & 
\multicolumn{1}{c}{\textcolor{green}{False}} & 
\\\midrule
\rotatebox{90}{State - After} & \multicolumn{1}{c}{\includegraphics[width=0.3\linewidth]{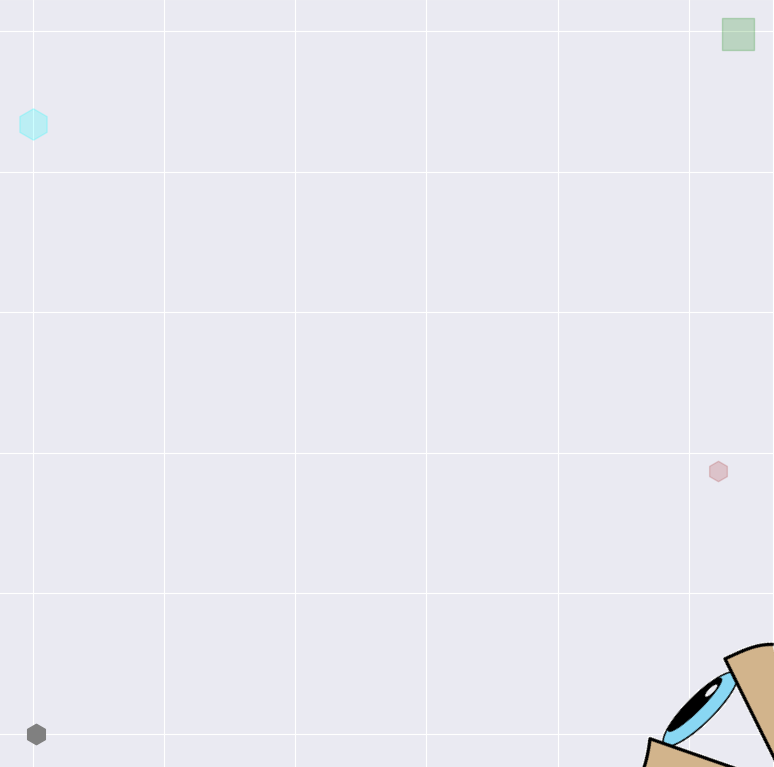}} &
\includegraphics[width=0.3\linewidth]{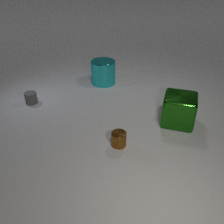} &
\\
 & \multicolumn{1}{c}{\textcolor{red}{}} & 
\multicolumn{1}{c}{\textcolor{red}{True}} & 
\\\bottomrule
\end{tabular}
\end{adjustbox}
\caption{Qualitative results on \textit{state-input} \vqaagent. For each example, a top-down view of the scene is presented. The eye at the bottom right of each view represents the rendering camera. It corresponds to the \vqaagent's viewpoint. Each scene is also rendered and presented side-by-side for comparison.}
\label{fig:state_vqa_manipulations}
\end{figure*}

\begin{figure*}[!htpb]
\begin{adjustbox}{max width=\linewidth}
\begin{tabular}{cccll}
\toprule
& \multicolumn{1}{c}{\large How many other things are both behind the purple rubber sphere \par} \\
& \multicolumn{1}{c}{\large and to the right of the gray cylinder? \par} \\
\rotatebox{90}{State - Before} & 
\includegraphics[width=0.3\linewidth]{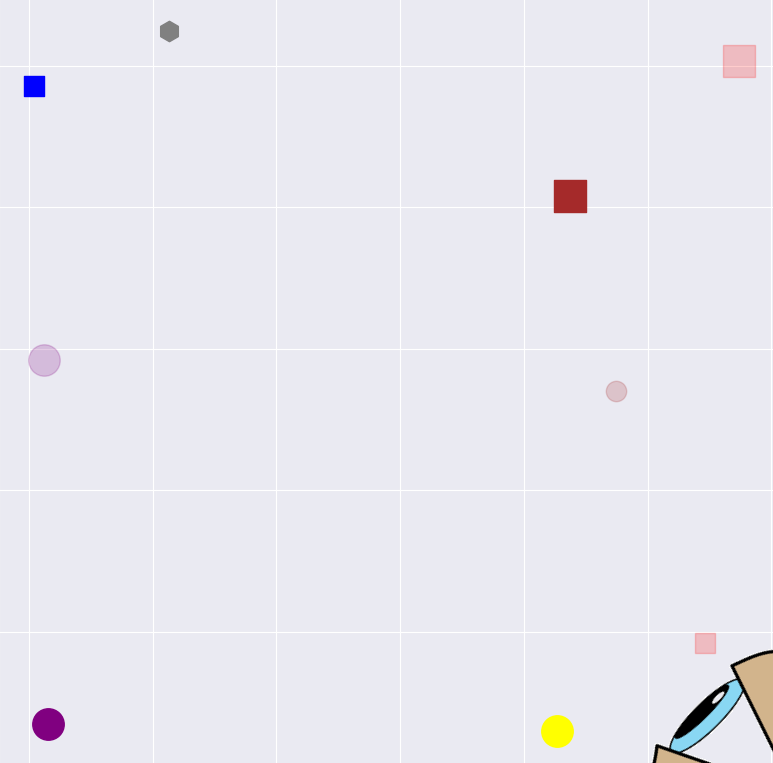} &
\includegraphics[width=0.3\linewidth]{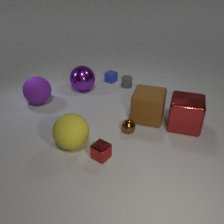}
\\
 & 
\multicolumn{1}{c}{\textcolor{green}{}} &
\multicolumn{1}{c}{\textcolor{green}{0}}
\\\midrule
\rotatebox{90}{State - After} & \includegraphics[width=0.3\linewidth]{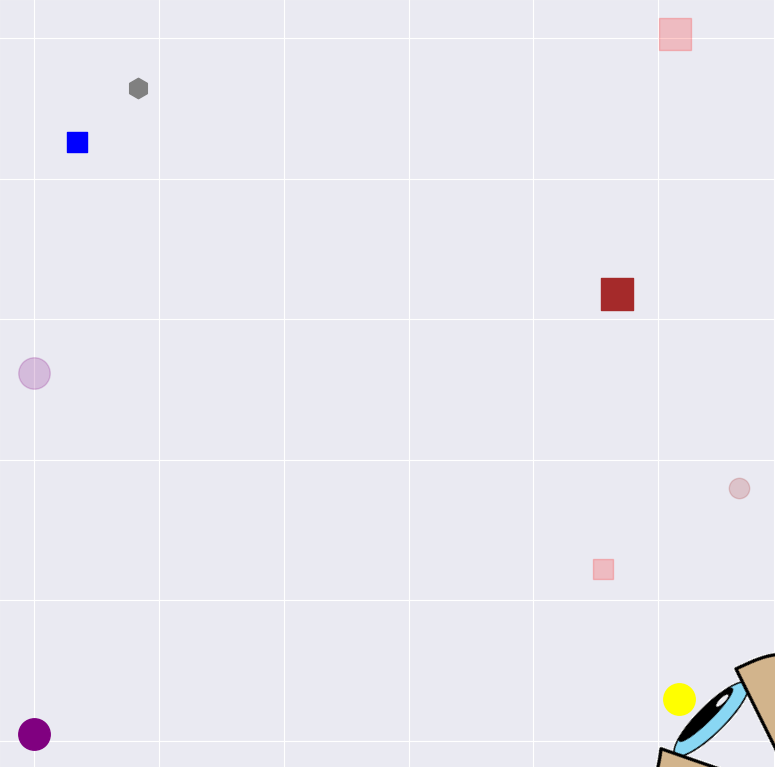} & \includegraphics[width=0.3\linewidth]{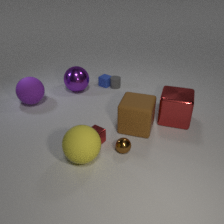}
\\
& \multicolumn{1}{c}{\textcolor{red}{}} &
\multicolumn{1}{c}{\textcolor{red}{1}}
\\
& \multicolumn{1}{c}{\large What is the material of the cube \par} \\
& \multicolumn{1}{c}{\large that is behind the rubber cube? \par} \\
\rotatebox{90}{State - Before} & \includegraphics[width=0.3\linewidth]{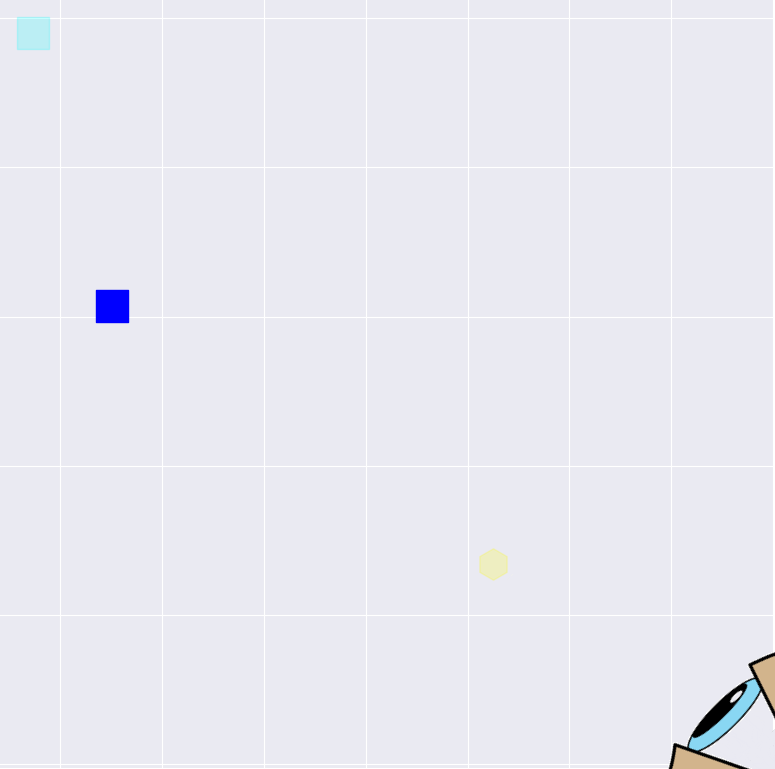} &
\includegraphics[width=0.3\linewidth]{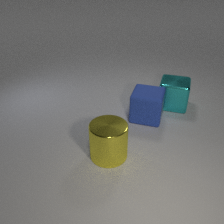}
\\
& \multicolumn{1}{c}{\textcolor{green}{}} &
\multicolumn{1}{c}{\textcolor{green}{Metal}}
\\\midrule
\rotatebox{90}{State - After} & \multicolumn{1}{c}{
\includegraphics[width=0.3\linewidth]{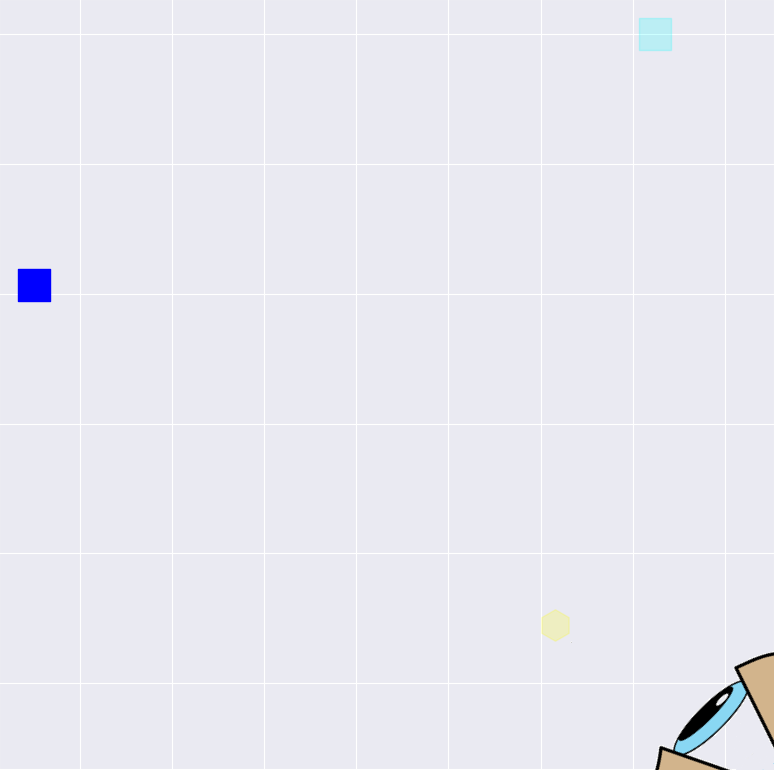}} &
\includegraphics[width=0.3\linewidth]{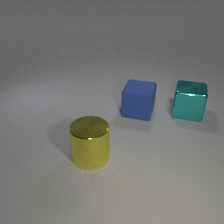}
\\
& \multicolumn{1}{c}{\textcolor{red}{}} &
\multicolumn{1}{c}{\textcolor{red}{Rubber}}
\\\bottomrule
\end{tabular}
\end{adjustbox}

\caption{Qualitative results on \textit{state-input} \vqaagent. For each example, a top-down view of the scene is presented. The eye at the bottom right of each view represents the rendering camera. It corresponds to the \vqaagent's viewpoint. Each scene is also rendered and presented side-by-side for comparison.}
\label{fig:state_vqa_manipulations_2}
\end{figure*}

Figures~\ref{fig:state_vqa_manipulations} and~\ref{fig:state_vqa_manipulations_2} show manipulations of \vqaagent that is trained to take \textit{states} instead of \textit{pixels} as the input. Such a \textit{state-input} \vqaagent receives a direct overview of the scene, bypassing any need for any image renderings. We use \textit{State-Input Transformer} (Section~\ref{sec:state_input_transformer}) as the \textit{multi-modal} component of that player. Since such a model gets the perfect visual information as the input, it makes it more robust under scene manipulations. As a consequence, our \adversarial tends to manipulate the scene so that objects are placed closely together. 

\clearpage
\subsection{Algorithms}
We show pseudo-algorithms that we use to (Algorithm~\ref{alg:dcl}) calculate rewards, (Algorithm~\ref{alg:tp}) train \adversarial, (Algorithm~\ref{alg:game}) and play a game.
\label{sec:algorithms}
\begin{algorithm}[!ht]
\caption{Calculate Rewards}\label{alg:dcl}
\begin{algorithmic}[1]

\State {$new\_answer:$ Answer produced by Visual Agent on perturbed image}
\State {$old\_answer:$ Answer produced by Visual Agent on original image}
\State {$gt\_answer:$  Ground Truth Answer}
\State {$dr:$ Drop Reward}
\State {$cr:$ Consistency Reward}
\State {$fr:$ Fail Reward}
\State $reward \gets 0$
\If {$new\_answer$ $\neq$ $old\_answer$}
    \If {$old\_answer$ $\neq$ $gt\_answer$}
        \State {$reward =  reward + cr$}
    \Else
        \State {$reward = reward + dr$}
    \EndIf
\Else
    \State {$reward = reward + fr$}
\EndIf
\State \textbf{return} $reward$
\end{algorithmic}
\end{algorithm}

\begin{algorithm}[!ht]
\caption{Training Pipeline}\label{alg:tp}
\begin{algorithmic}[1]

\State {$M:$ Minigames} 
\State {$Va:$ Visual Agent, $Fe:$ Feature Extractor}
\State {$A:$ Actor, $C:$ Critic}
\State {$Qre:$ Question Relevance Enforcer, $Sce:$ Scene Constraint Enforcer}

\For{$batch$ $in$ $M$}
    \State {$rewards, state\_values, logprobs \gets Data\space Game(M,Va,Fe,A,C,Qre,Sce)$}
    \State $n \gets |rewards|$
    \State {$advantage \gets rewards - state\_values$}
    \State {$ploss \gets -logprobs \times advantage $}
    \State {$vloss \gets (state\_values - stop\_grad(rewards))^2 / n$}
    \State {$loss \gets ploss + vloss$}
    \State {$backprop(loss, A, C)$}
\EndFor
\State \textbf{return}
\end{algorithmic}
\end{algorithm}

\begin{algorithm}[!ht]
\caption{Game between players (Game)}\label{alg:game}
\begin{algorithmic}[1]

\State {$M:$ \minigame}
\State {$\mathcal{P}_{\text{VQA}}:$ Visual Agent, $\mathcal{S}_k:$ Multi-modal Backbone}
\State {$A:$ Actor, $C:$ Critic} 
\State {$Qre:$ Question Relevance Enforcer, $Sce:$ Scene Constraint Enforcer}
\State {$isr:$ Invalid Scene Reward}
\For{$batch$ $in$ $M$}
    \State $rewards \gets \{\}$
    \State $state\_values \gets \{\}$
    \State $logprobs \gets \{\}$
    \For{$(image, scene, question, program, gt\_answer)$ $in$ $batch$}
        \State {$old\_answer \gets \mathcal{P}_{\text{VQA}}(image, question)$}
        \State {$extracted\_features \gets \mathcal{S}_k(scene, question)$}
        \State {$dx\_probs, dy\_probs \gets A(extracted\_features)$}
        \State {$CatX\gets Categorical(dx\_probs)$}
        \State {$CatY\gets Categorical(dy\_probs)$}
        \State {$state\_value \gets C(extracted\_features)$}
        \State {$dx \gets $ sample $CatX$}
        \State {$dy \gets $ sample $CatY$}
        \State {$new\_scene \gets Perturbate(scene, dx, dy)$}
        \State {$check\_scene \gets Sce(new\_scene)$}
        \State {$check\_question \gets Qre(new\_scene,program,gt\_answer)$}
        \If {$check\_scene \land check\_question =\ True$}
            \State $new\_image \gets Render(new\_scene)$
            \State $new\_answer \gets \mathcal{P}_{\text{VQA}}(new\_image, question)$
            \State {$rewards\gets rewards\, ||\, CalcRewards(new\_answer, old\_answer,gt\_answer)$}
        \Else
            \State {$rewards\gets rewards\, ||\, isr$}
        \EndIf
        \State {$state\_values\gets state\_values\, ||\, state\_value$}
        \State {$logprob \gets logprob ||\, log(p(dx|CatX))) + log(p(dy|CatY)))$}
    \EndFor
\EndFor
\State \textbf{return} $rewards, state\_values, logprobs$
\end{algorithmic}
\end{algorithm}

\clearpage
\subsection{Sensitivity to Question Types}
\label{sec:sensitivity_question_types}
\begin{figure}[!ht]
     \centering
     \begin{subfigure}[b]{0.49\textwidth}
         \centering
         \includegraphics[width=0.8\textwidth]{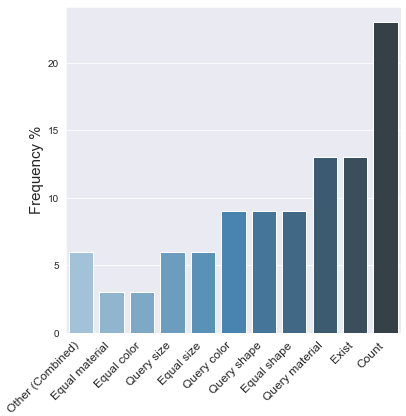}
         \caption{FiLM}
     \end{subfigure}
     \hfill
     \begin{subfigure}[b]{0.49\textwidth}
         \centering
         \includegraphics[width=0.8\textwidth]{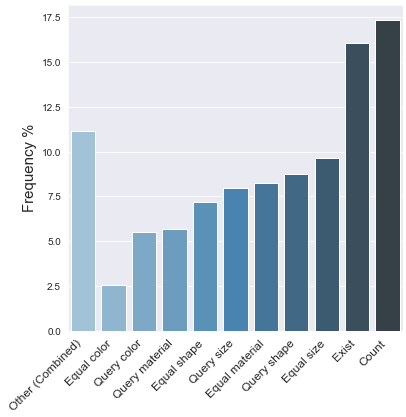}
         \caption{TbD}
     \end{subfigure}
     \begin{subfigure}[b]{\textwidth}
         \centering
         \includegraphics[width=0.4\textwidth]{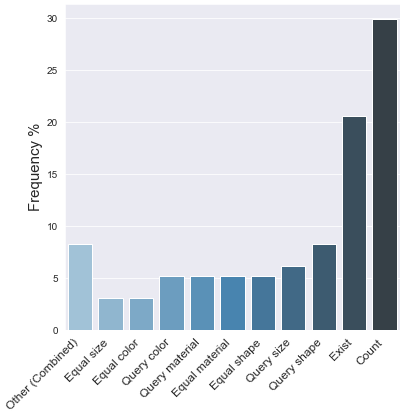}
         \caption{Mdetr}
     \end{subfigure}
     \caption{Histograms of the model's susceptibility to scene manipulations conditioned on a question type.}
     \label{fig:question_types}
\end{figure}

Figure~\ref{fig:question_types} shows the susceptibility of CLEVR models to scene manipulations of \adversarial. We observe that especially \textit{counting} and \textit{existence} questions are the `back-doors` for our scene manipulations. Questions of those two types typically involve multiple steps of reasoning. \adversarial's scene manipulations make them more likely that reasoning will fail at some stage.

\clearpage

\subsection{Convergence}
\label{sec:convergence}
\begin{figure*}[!ht]
\begin{center}
\begin{tabular}{c@{\ }c@{\ }c@{\ }c@{\ }c@{\ }c@{\ }c}
& \multicolumn{1}{c}{\minigame size 100} & 
\multicolumn{1}{c}{\minigame size 1000} \\
\rotatebox{90}{\quad\quad\quad Film} &
\includegraphics[width=0.41\linewidth]{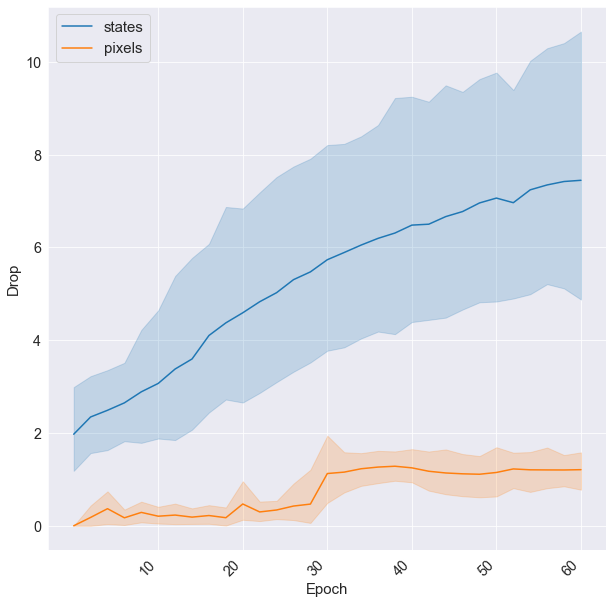} & 
\includegraphics[width=0.41\linewidth]{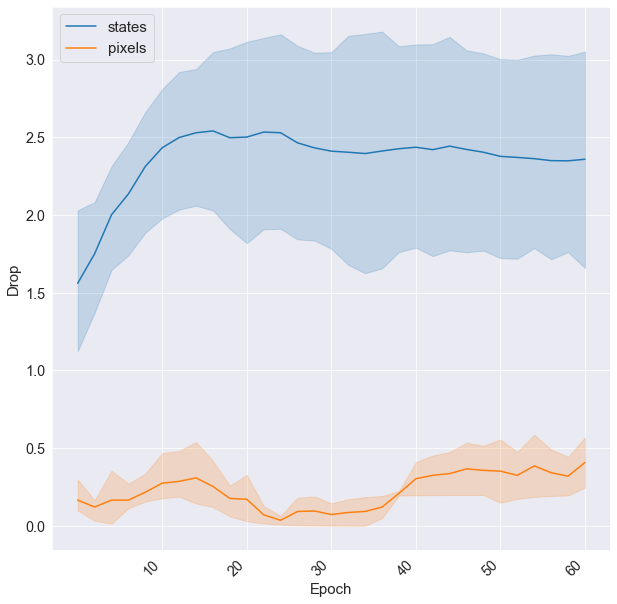} &
\\
\rotatebox{90}{\quad\quad\quad TbD} &
\includegraphics[width=0.41\linewidth]{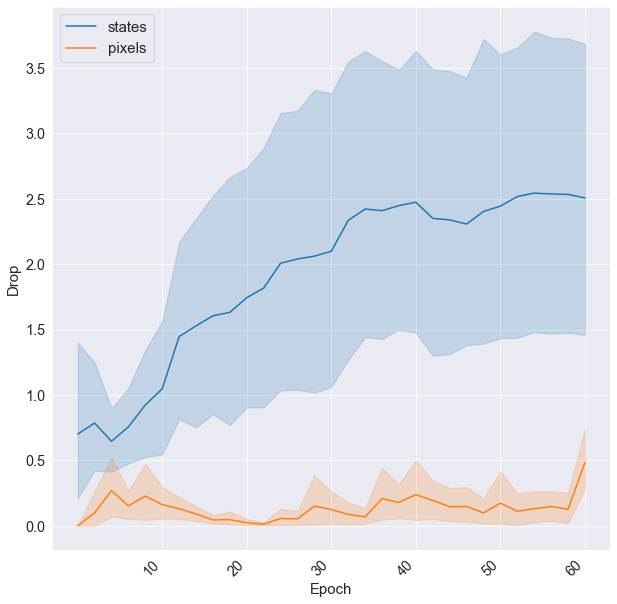} & 
\includegraphics[width=0.41\linewidth]{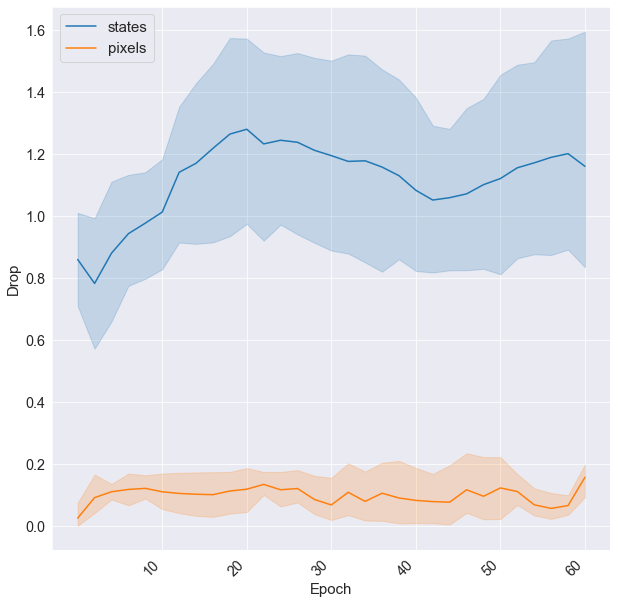} &
\\
\rotatebox{90}{\quad\quad\quad MDetr} &
\includegraphics[width=0.41\linewidth]{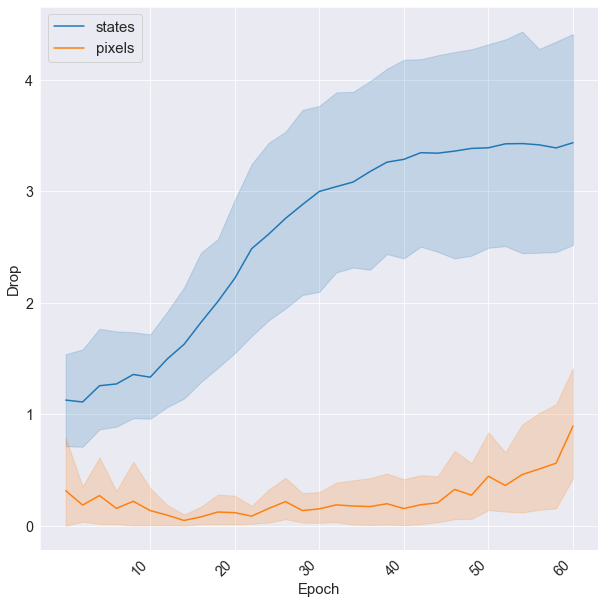} & 
\includegraphics[width=0.41\linewidth]{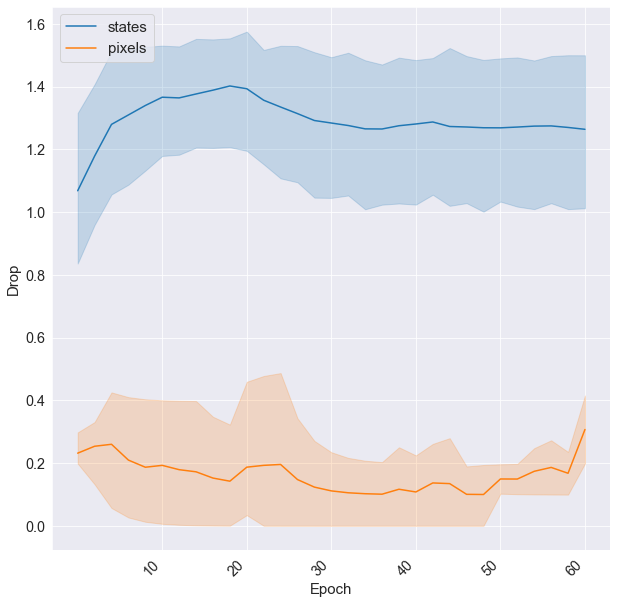} &

\end{tabular}
\end{center}
\caption{Performance of \adversarial for different training snapshots. We show the mean and variance over thirty trials. \adversarial is either trained from \textit{states} (blue) or \textit{pixels} (orange).}
\label{agent_performance}
\end{figure*}

Figure~\ref{agent_performance} shows convergence plots of \adversarial during training against FiLM, TbD and MDetr. We use different initialization seeds and $30$ trials. \adversarial is trained either from \textit{states} or \textit{pixels}. The former yields higher performance and sometimes better convergence. We use the \textit{Drop} metric that measures the accuracy drop after the \vqaagent is manipulated. The higher \textit{Drop}, the more successful manipulations are.

\clearpage
\subsection{Performance Distribution Plots}
\label{sec:performance_distribution}
\begin{figure*}[!ht]
\begin{center}
\begin{tabular}{c@{\ }c@{\ }c@{\ }c@{\ }c@{\ }c@{\ }c}
& \multicolumn{1}{c}{\minigame size 100} & 
\multicolumn{1}{c}{\minigame size 1000} \\
\rotatebox{90}{\quad\quad\quad Film} &
\includegraphics[width=0.41\linewidth]{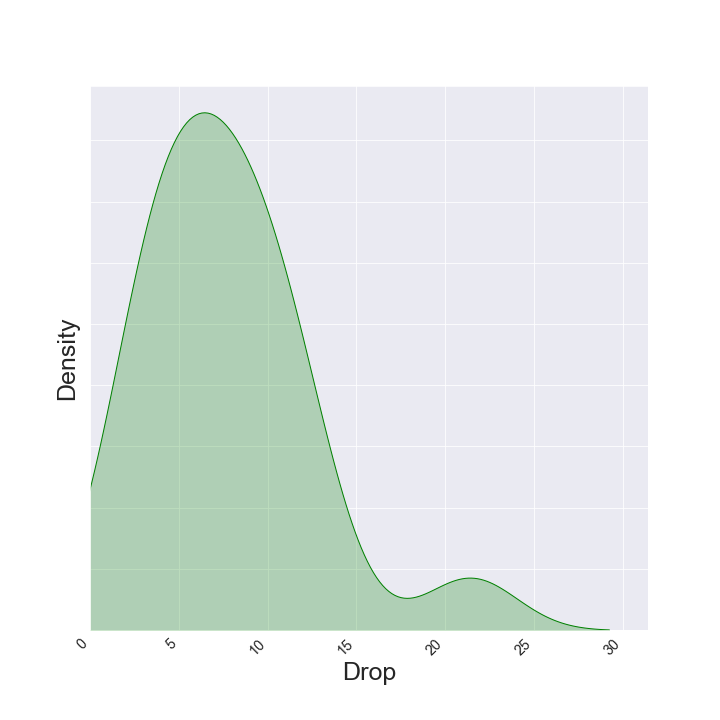} & 
\includegraphics[width=0.41\linewidth]{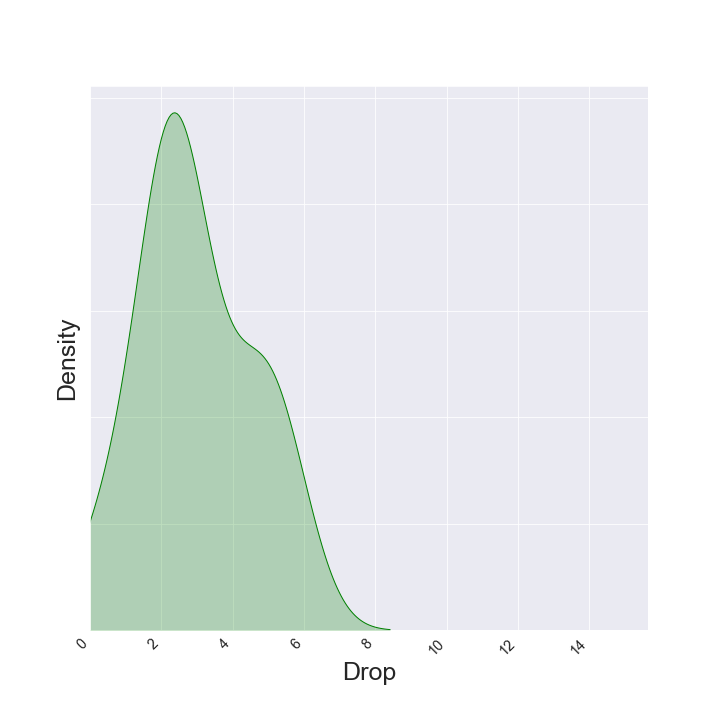} &
\\
\rotatebox{90}{\quad\quad\quad TbD} &
\includegraphics[width=0.41\linewidth]{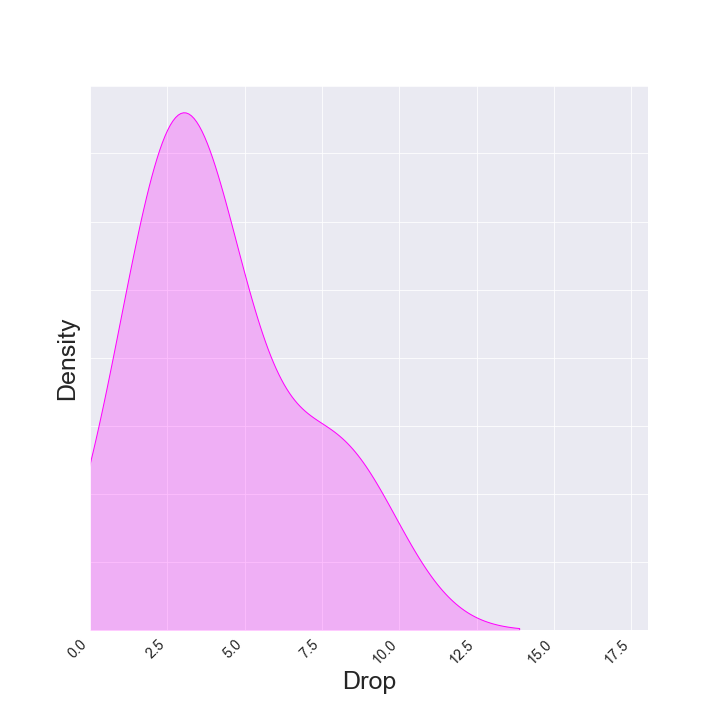} & 
\includegraphics[width=0.41\linewidth]{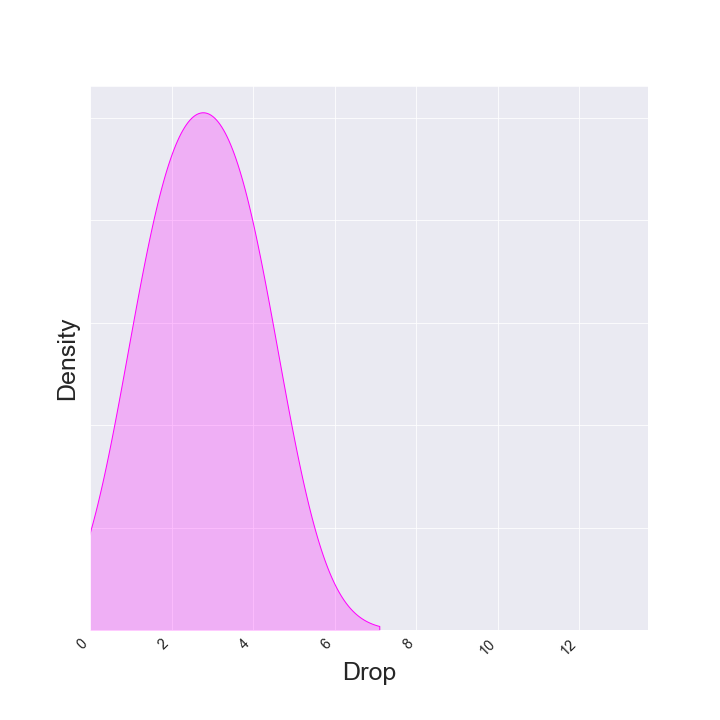} &
\\
\rotatebox{90}{\quad\quad\quad MDetr} &
\includegraphics[width=0.41\linewidth]{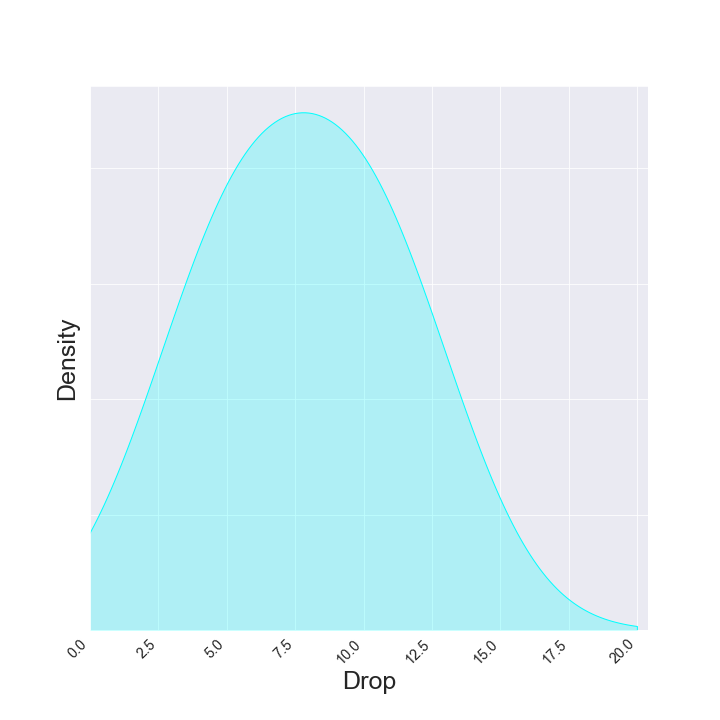} & 
\includegraphics[width=0.41\linewidth]{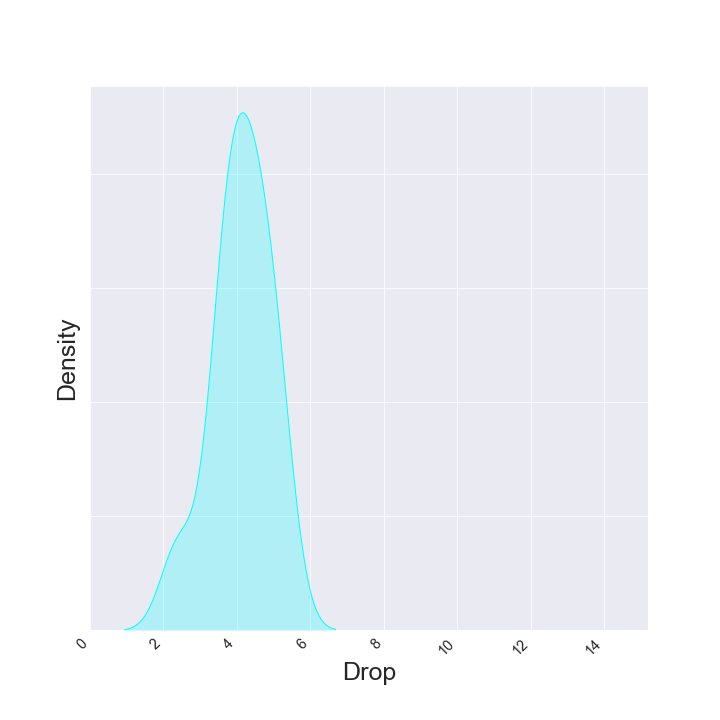} &

\end{tabular}
\end{center}
\caption{\minigames performance distribution for \textit{state-input} \adversarial.}
\label{state_agent_performance_dist}
\end{figure*}

\begin{figure*}[!ht]
\begin{center}
\begin{tabular}{c@{\ }c@{\ }c@{\ }c@{\ }c@{\ }c@{\ }c}
& \multicolumn{1}{c}{\minigame size 100} & 
\multicolumn{1}{c}{\minigame size 1000} \\
\rotatebox{90}{\quad\quad\quad Film } &
\includegraphics[width=0.45\linewidth]{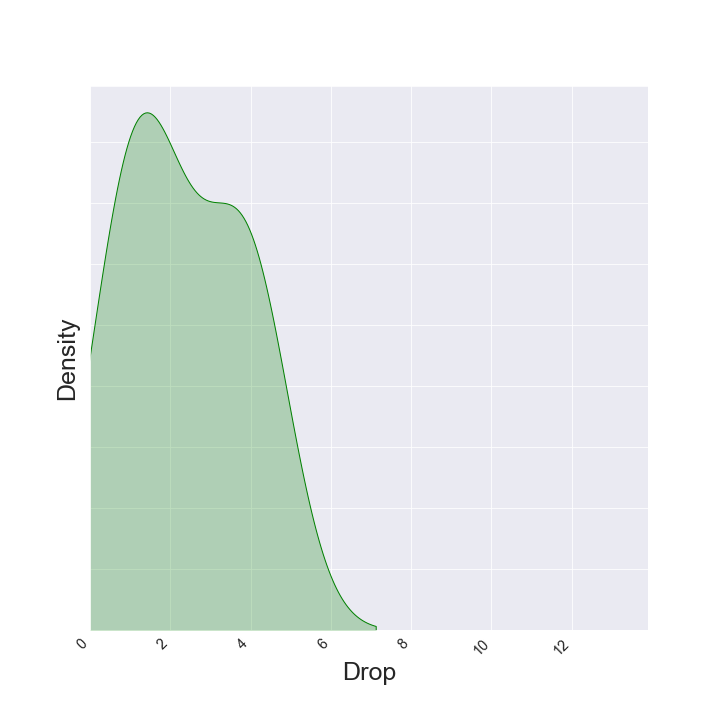} & 
\includegraphics[width=0.45\linewidth]{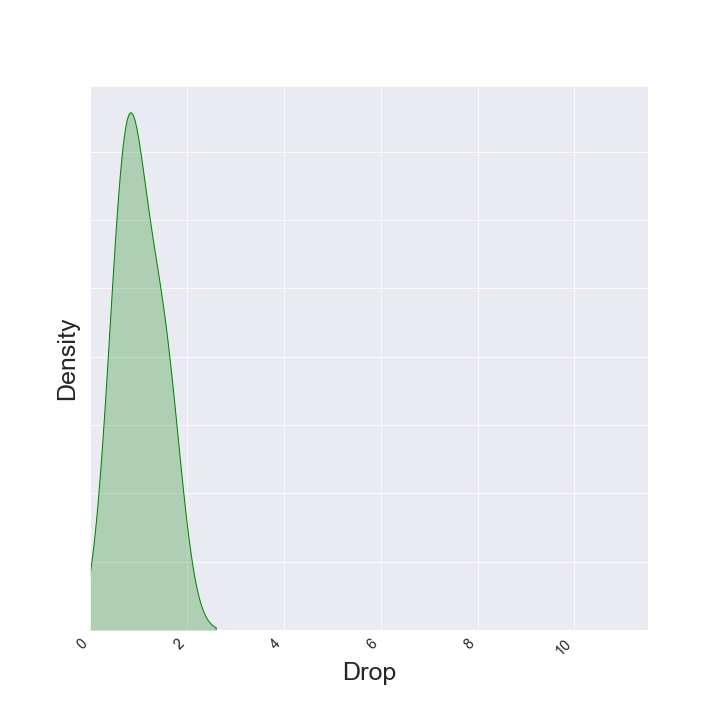} &
\\
\rotatebox{90}{\quad\quad\quad TbD } &
\includegraphics[width=0.45\linewidth]{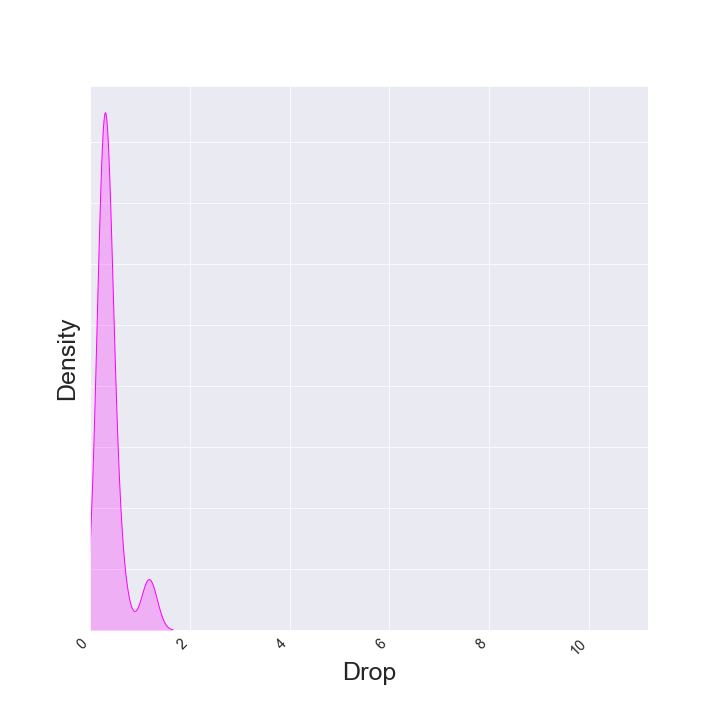} & 
\includegraphics[width=0.45\linewidth]{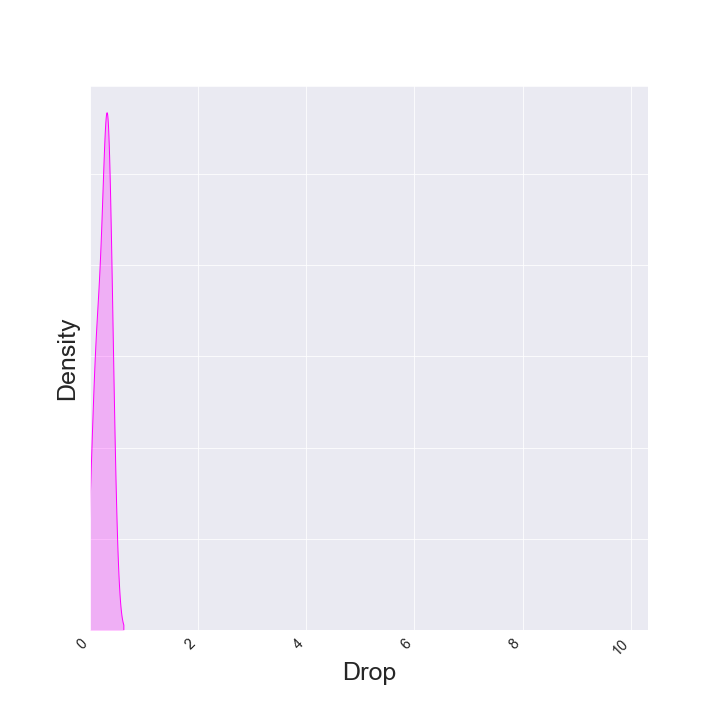} &
\\
\rotatebox{90}{\quad\quad\quad MDetr } &
\includegraphics[width=0.45\linewidth]{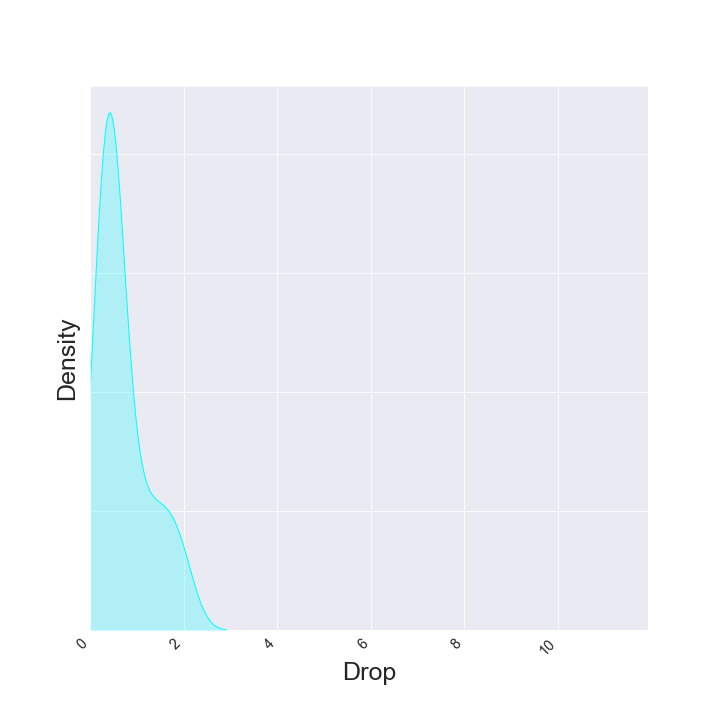} & 
\includegraphics[width=0.45\linewidth]{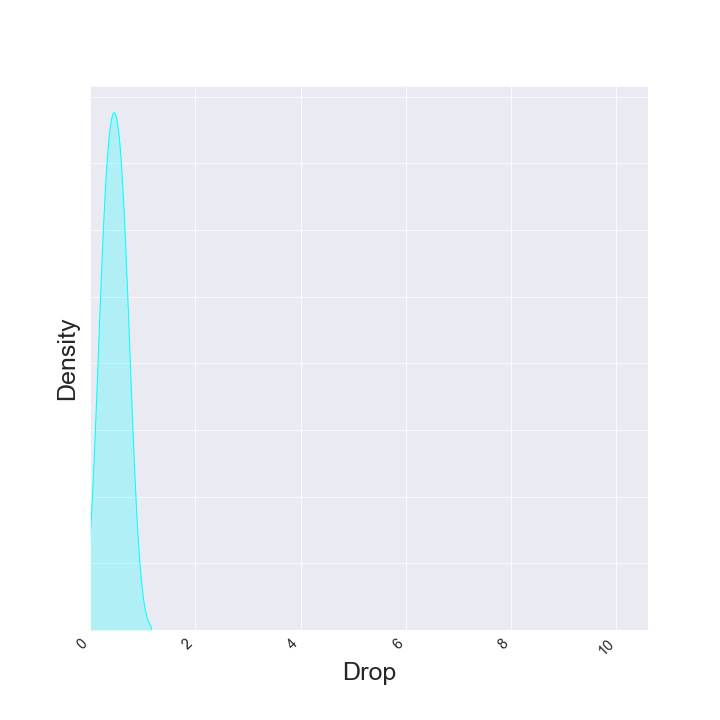} &

\end{tabular}
\end{center}
\caption{\minigames performance distribution for \textit{pixel-input} \adversarial.}
\label{image_agent_performance_dist}
\end{figure*}

Figures~\ref{state_agent_performance_dist} and~\ref{image_agent_performance_dist} show the distribution of the \adversarial performance, where we show how often (y-axis) the given \textit{Drop} score is achieved (x-axis). The \textit{Drop} metric measures the accuracy drop after the \vqaagent is manipulated. The higher \textit{Drop}, the more successful manipulations are. Larger \minigames lead to more consistent (narrower distributions) but lower performance. This behavior suggests that scene diversity present in larger \minigames leads to more stable scene manipulations. However, they do so at the cost of increasing the complexity of the optimization problem.

\clearpage
\subsection{Out-of-Distribution Manipulations}
\label{sec:stretching_behavior}
\begin{figure}[!ht]
     \centering
     \begin{subfigure}[b]{0.49\textwidth}
         \centering
         \includegraphics[width=0.8\textwidth]{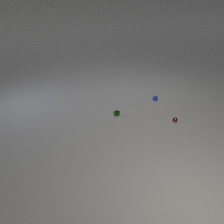}
         \caption{ Shape of the blue cube is misclassified as sphere.}
     \end{subfigure}
     \hfill
     \begin{subfigure}[b]{0.49\textwidth}
         \centering
         \includegraphics[width=0.8\textwidth]{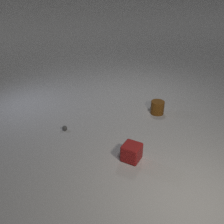}   
         \caption{ Large, brown cylinder is misclassified as small.}
     \end{subfigure}
     \hfill
     \begin{subfigure}[b]{0.49\textwidth}
         \centering
         \includegraphics[width=0.8\textwidth]{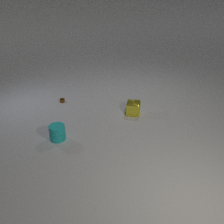}   
         \caption{ Large, yellow cube is misclassified as small.}
     \end{subfigure}
     \caption{Out-of-distribution examples.}
     \label{fig:out_distribution}
\end{figure}

If we do not provide extra constraints on the scene generation process with our environment enforcers, \adversarial may find an easy manipulation that changes the camera pose or location, effectively `zooming out' the whole scene. It does so by stretching all the object coordinates in the scene. Figure~\ref{fig:out_distribution} illustrates that. For instance, \textit{large} objects look much smaller than typical examples. As such constructed scenes are quite different from the ones that \vqaagent has observed during training, we categorize such manipulations to be \textit{out-of-distribution}. Note that, due to our \textit{in-distribution} environment enforcers, such scenes are prohibited in our pipeline.

\clearpage
\subsection{Additional Qualitative Results}
\label{sec:additional_examples}
\begin{figure*}[!htpb]
\begin{adjustbox}{max width=\textwidth}
\begin{tabular}{l@{\ }l@{\ }c@{\ }c@{\ }c@{\ }c}
 \toprule
 & \multicolumn{1}{c}{\huge Is the number of small cubes that \par} & 
\multicolumn{1}{c}{\huge What is the color of the metallic thing \par } \\
& \multicolumn{1}{c}{\huge are right of the blue rubber block greater  \par} & 
\multicolumn{1}{c}{\huge that is on the right side of \par} \\
& \multicolumn{1}{c}{\huge than the number of small balls that\par} & 
\multicolumn{1}{c}{\huge the shiny thing behind \par}\\
& \multicolumn{1}{c}{\huge  are in front of the cyan object?\par} & 
\multicolumn{1}{c}{\huge the large gray object?}
\\\midrule
\rotatebox{90}{\Huge SAN - Before \par} & \multicolumn{1}{c}{\includegraphics[width=0.8\linewidth]{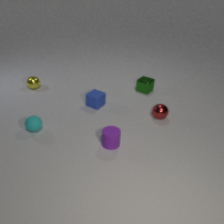}} &
\includegraphics[width=0.8\linewidth]{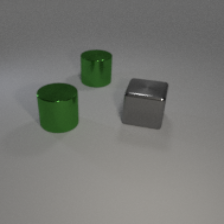}
\\
 & \multicolumn{1}{c}{{\Huge \textcolor{green}{Yes} \par}} & 
\multicolumn{1}{c}{\Huge \textcolor{green}{Gray} \par}
\\\midrule
\rotatebox{90}{\Huge SAN - After \par} & \multicolumn{1}{c}{\includegraphics[width=0.8\linewidth]{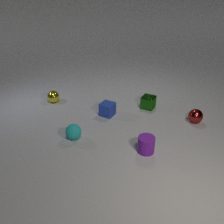}} &
\includegraphics[width=0.8\linewidth]{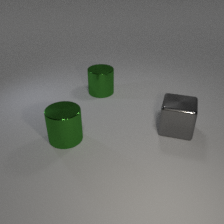}
\\
 & \multicolumn{1}{c}{{\Huge \textcolor{red}{No} \par}} & 
\multicolumn{1}{c}{\Huge \textcolor{red}{Green} \par}
\\\bottomrule
\end{tabular}
\end{adjustbox}

\caption{Manipulations of CLEVR models. We show results \textit{before} and \textit{after} scene manipulations.}
\label{fig:san_qualitative}
\end{figure*}

\begin{figure*}[!htpb]
\begin{adjustbox}{width=\textwidth}
\begin{tabular}{l@{\ }l@{\ }c@{\ }c@{\ }c@{\ }c}
 \toprule
 & \multicolumn{1}{c}{\huge There is a cube on the left\par} & 
\multicolumn{1}{c}{\huge What shape is the thing that \par } \\
& \multicolumn{1}{c}{\huge side of the brown object \par} & 
\multicolumn{1}{c}{\huge is in front of the small yellow  \par} \\
& \multicolumn{1}{c}{\huge that is behind the red ball;\par} & 
\multicolumn{1}{c}{\huge cylinder and to the right \par}\\
& \multicolumn{1}{c}{\huge what size is it? \par} & 
\multicolumn{1}{c}{\huge of the rubber thing? \par}
\\\midrule
\rotatebox{90}{\Huge SAN - Before \par} & \multicolumn{1}{c}{\includegraphics[width=0.7\linewidth]{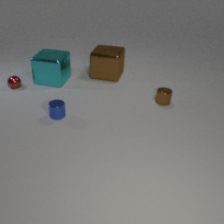}} &
\includegraphics[width=0.7\linewidth]{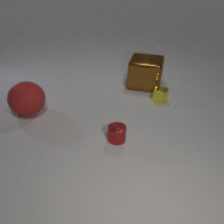}
\\
 & \multicolumn{1}{c}{{\Huge \textcolor{green}{Large} \par}} & 
\multicolumn{1}{c}{\Huge \textcolor{green}{Cylinder} \par}
\\\midrule
\rotatebox{90}{\Huge SAN - After \par} & \multicolumn{1}{c}{\includegraphics[width=0.7\linewidth]{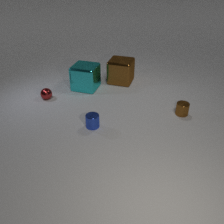}} &
\includegraphics[width=0.7\linewidth]{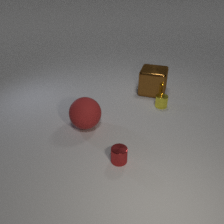}
\\
 & \multicolumn{1}{c}{{\Huge \textcolor{red}{Small} \par}} & 
\multicolumn{1}{c}{\Huge \textcolor{red}{Sphere} \par}
\\\bottomrule
\end{tabular}
\end{adjustbox}

\caption{Manipulations of CLEVR models. We show results \textit{before} and \textit{after} scene manipulations.}
\label{fig:san_qualitative_2}
\end{figure*}

\begin{figure*}[!htpb]
\begin{adjustbox}{max width=\textwidth}
\begin{tabular}{l@{\ }l@{\ }c@{\ }c@{\ }c@{\ }c}
 \toprule
 & \multicolumn{1}{c}{\huge How big is the shiny cylinder \par} & 
\multicolumn{1}{c}{\huge There is a thing that is the\par } \\
& \multicolumn{1}{c}{\huge that is behind the \par} & 
\multicolumn{1}{c}{\huge same color as the cube;\par} \\
& \multicolumn{1}{c}{\huge rubber cylinder? \par} & 
\multicolumn{1}{c}{\huge  what is it made of?\par}\\
& \multicolumn{1}{c}{\huge  \par} & 
\multicolumn{1}{c}{\huge  \par}
\\\midrule
\rotatebox{90}{\Huge IEP - Before \par} & \multicolumn{1}{c}{\includegraphics[width=0.7\linewidth]{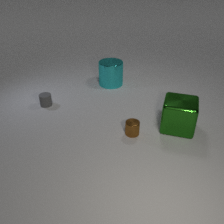}} &
\includegraphics[width=0.7\linewidth]{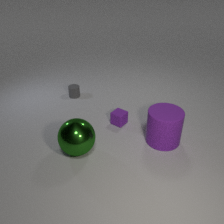}
\\
 & \multicolumn{1}{c}{{\Huge \textcolor{green}{Large} \par}} & 
\multicolumn{1}{c}{\Huge \textcolor{green}{Rubber} \par}
\\\midrule
\rotatebox{90}{\Huge IEP - After \par} & \multicolumn{1}{c}{\includegraphics[width=0.7\linewidth]{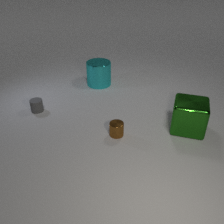}} &
\includegraphics[width=0.7\linewidth]{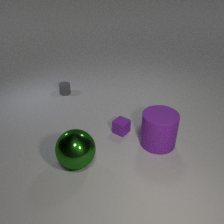}
\\
 & \multicolumn{1}{c}{{\Huge \textcolor{red}{Small} \par}} & 
\multicolumn{1}{c}{\Huge \textcolor{red}{Metal} \par}
\\\bottomrule
\end{tabular}
\end{adjustbox}

\caption{Manipulations of CLEVR models. We show results \textit{before} and \textit{after} scene manipulations.}
\label{fig:iep_qualitative}
\end{figure*}

\begin{figure*}[!htpb]
\begin{adjustbox}{max width=\textwidth}
\begin{tabular}{l@{\ }l@{\ }c@{\ }c@{\ }c@{\ }c}
 \toprule
 & \multicolumn{1}{c}{\huge Does the cyan metal object that \par} & 
\multicolumn{1}{c}{\huge What is the big thing that is behind \par } \\
& \multicolumn{1}{c}{\huge is behind the tiny gray block have \par} & 
\multicolumn{1}{c}{\huge the large red shiny sphere \par} \\
& \multicolumn{1}{c}{\huge the same size as the block behind \par} & 
\multicolumn{1}{c}{\huge  made of?\par}\\
& \multicolumn{1}{c}{\huge  the small cyan metallic objec?\par} & 
\multicolumn{1}{c}{\huge  \par}
\\\midrule
\rotatebox{90}{\Huge IEP - Before \par} & \multicolumn{1}{c}{\includegraphics[width=0.7\linewidth]{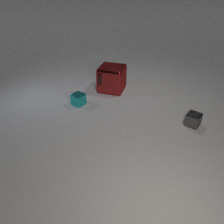}} &
\includegraphics[width=0.7\linewidth]{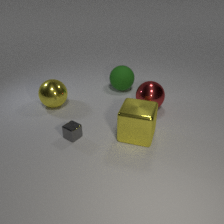}
\\
 & \multicolumn{1}{c}{{\Huge \textcolor{green}{No} \par}} & 
\multicolumn{1}{c}{\Huge \textcolor{green}{Rubber} \par}
\\\midrule
\rotatebox{90}{\Huge IEP - After \par} & \multicolumn{1}{c}{\includegraphics[width=0.7\linewidth]{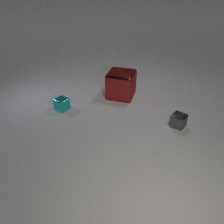}} &
\includegraphics[width=0.7\linewidth]{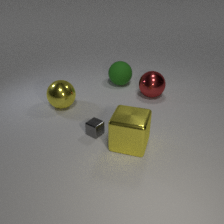}
\\
 & \multicolumn{1}{c}{{\Huge \textcolor{red}{Yes} \par}} & 
\multicolumn{1}{c}{\Huge \textcolor{red}{Metal} \par}
\\\bottomrule
\end{tabular}
\end{adjustbox}

\caption{Manipulations of CLEVR models. We show results \textit{before} and \textit{after} scene manipulations.}
\label{fig:iep_qualitative_2}
\end{figure*}

\begin{figure*}[!htpb]
\begin{adjustbox}{max width=\textwidth}
\begin{tabular}{l@{\ }l@{\ }c@{\ }c@{\ }c@{\ }c}
 \toprule
 & \multicolumn{1}{c}{\huge There is a metallic object; \par} & 
\multicolumn{1}{c}{\huge How many other small shiny objects\par } \\
& \multicolumn{1}{c}{\huge does it have the same size as\par} & 
\multicolumn{1}{c}{\huge are the same shape as the \par} \\
& \multicolumn{1}{c}{\huge the rubber block right of the\par} & 
\multicolumn{1}{c}{\huge  red object?\par}\\
& \multicolumn{1}{c}{\huge  tiny shiny cube?\par} & 
\multicolumn{1}{c}{}
\\\midrule
\rotatebox{90}{\Huge RN - Before \par} & \multicolumn{1}{c}{\includegraphics[width=0.7\linewidth]{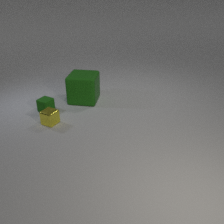}} &
\includegraphics[width=0.7\linewidth]{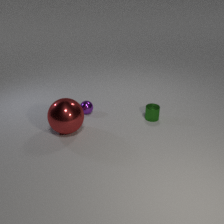}
\\
 & \multicolumn{1}{c}{{\Huge \textcolor{green}{No} \par}} & 
\multicolumn{1}{c}{\Huge \textcolor{green}{1} \par}
\\\midrule
\rotatebox{90}{\Huge RN - After \par} & \multicolumn{1}{c}{\includegraphics[width=0.7\linewidth]{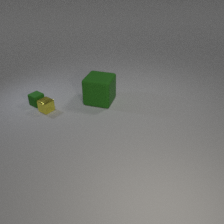}} &
\includegraphics[width=0.7\linewidth]{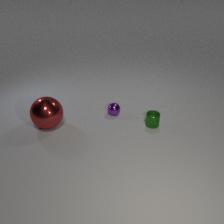}
\\
 & \multicolumn{1}{c}{{\Huge \textcolor{red}{Yes} \par}} & 
\multicolumn{1}{c}{\Huge \textcolor{red}{2} \par}
\\\bottomrule
\end{tabular}
\end{adjustbox}
\caption{Manipulations of CLEVR models. We show results \textit{before} and \textit{after} scene manipulations.}
\label{fig:rn_qualitative}
\end{figure*}

\begin{figure*}[!htpb]
\begin{adjustbox}{max width=\textwidth}
\begin{tabular}{l@{\ }l@{\ }c@{\ }c@{\ }c@{\ }c}
 \toprule
 & \multicolumn{1}{c}{\huge  \par} & 
\multicolumn{1}{c}{\huge Are there any other things  \par } \\
& \multicolumn{1}{c}{\huge Are there any other things \par} & 
\multicolumn{1}{c}{\huge that are the same shape as the\par} \\
& \multicolumn{1}{c}{\huge that are the same size\par} & 
\multicolumn{1}{c}{\huge big metallic object? \par}\\
& \multicolumn{1}{c}{\huge  as the ball?\par} & 
\multicolumn{1}{c}{\huge  \par}
\\\midrule
\rotatebox{90}{\Huge RN - Before \par} & \multicolumn{1}{c}{\includegraphics[width=0.7\linewidth]{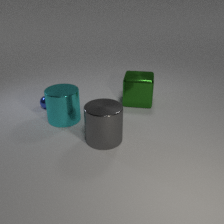}} &
\includegraphics[width=0.7\linewidth]{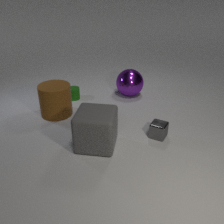}
\\
 & \multicolumn{1}{c}{{\Huge \textcolor{green}{No} \par}} & 
\multicolumn{1}{c}{\Huge \textcolor{green}{No} \par}
\\\midrule
\rotatebox{90}{\Huge  RN - After\par} & \multicolumn{1}{c}{\includegraphics[width=0.7\linewidth]{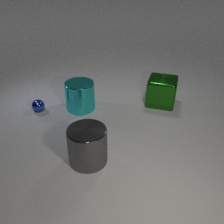}} &
\includegraphics[width=0.7\linewidth]{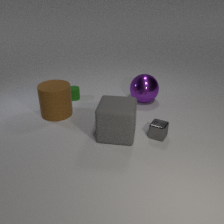}
\\
 & \multicolumn{1}{c}{{\Huge \textcolor{red}{Yes} \par}} & 
\multicolumn{1}{c}{\Huge \textcolor{red}{Yes} \par}
\\\bottomrule
\end{tabular}
\end{adjustbox}

\caption{Manipulations of CLEVR models. We show results \textit{before} and \textit{after} scene manipulations.}
\label{fig:rn_qualitative_2}
\end{figure*}

\begin{figure*}[!htpb]
\begin{adjustbox}{max width=\textwidth}
\begin{tabular}{l@{\ }l@{\ }c@{\ }c@{\ }c@{\ }c}
 \toprule
 & \multicolumn{1}{c}{\huge How many objects are there \par} & 
\multicolumn{1}{c}{\huge  There is another object that is the \par } \\
& \multicolumn{1}{c}{\huge of the same size as \par} & 
\multicolumn{1}{c}{\huge same size as the blue\par} \\
& \multicolumn{1}{c}{\huge the brown object?\par} & 
\multicolumn{1}{c}{\huge rubber object; What is it \par}\\
& \multicolumn{1}{c}{\huge  \par} & 
\multicolumn{1}{c}{\huge  made of?\par}
\\\midrule
\rotatebox{90}{\Huge MDetr - Before \par} & \multicolumn{1}{c}{\includegraphics[width=0.7\linewidth]{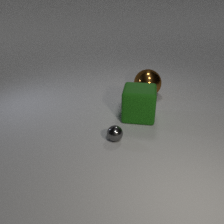}} &
\includegraphics[width=0.7\linewidth]{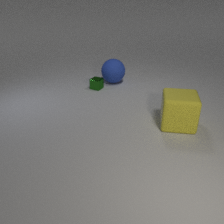}
\\
 & \multicolumn{1}{c}{{\Huge \textcolor{green}{1} \par}} & 
\multicolumn{1}{c}{\Huge \textcolor{green}{Rubber} \par}
\\\midrule
\rotatebox{90}{\Huge  MDetr - After\par} & \multicolumn{1}{c}{\includegraphics[width=0.7\linewidth]{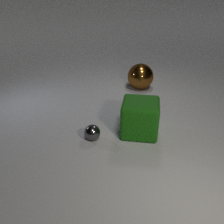}} &
\includegraphics[width=0.7\linewidth]{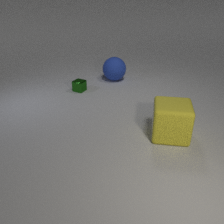}
\\
 & \multicolumn{1}{c}{{\Huge \textcolor{red}{0} \par}} & 
\multicolumn{1}{c}{\Huge \textcolor{red}{Metal} \par}
\\\bottomrule
\end{tabular}
\end{adjustbox}

\caption{Manipulations of CLEVR models. We show results \textit{before} and \textit{after} scene manipulations.}
\label{fig:mdetr_qualitative}
\end{figure*}

\begin{figure*}[!htpb]
\begin{adjustbox}{max width=\textwidth}
\begin{tabular}{l@{\ }l@{\ }c@{\ }c@{\ }c@{\ }c}
 \toprule
 & \multicolumn{1}{c}{\huge  What number of other objects\par} & 
\multicolumn{1}{c}{\huge   What number of metal objects\par } \\
& \multicolumn{1}{c}{\huge  are the same color \par} & 
\multicolumn{1}{c}{\huge are small objects or\par} \\
& \multicolumn{1}{c}{\huge as the block ?\par} & 
\multicolumn{1}{c}{\huge  big cyan objects?\par}\\
& \multicolumn{1}{c}{\huge  \par} & 
\multicolumn{1}{c}{\huge  \par}
\\\midrule
\rotatebox{90}{\Huge MDetr - Before \par} & \multicolumn{1}{c}{\includegraphics[width=0.7\linewidth]{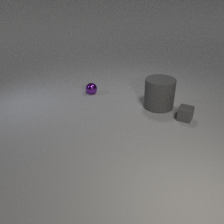}} &
\includegraphics[width=0.7\linewidth]{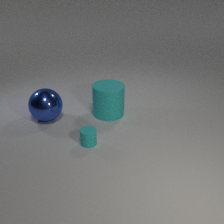}
\\
 & \multicolumn{1}{c}{{\Huge \textcolor{green}{1} \par}} & 
\multicolumn{1}{c}{\Huge \textcolor{green}{0} \par}
\\\midrule
\rotatebox{90}{\Huge  MDetr - After\par} & \multicolumn{1}{c}{\includegraphics[width=0.7\linewidth]{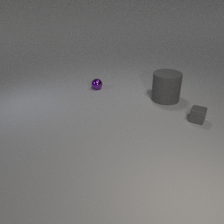}} &
\includegraphics[width=0.7\linewidth]{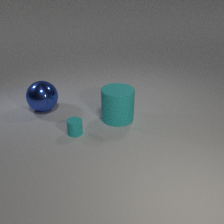}
\\
 & \multicolumn{1}{c}{{\Huge \textcolor{red}{0} \par}} & 
\multicolumn{1}{c}{\Huge \textcolor{red}{1} \par}
\\\bottomrule
\end{tabular}
\end{adjustbox}

\caption{Manipulations of CLEVR models. We show results \textit{before} and \textit{after} scene manipulations.}
\label{fig:mdetr_qualitative_2}
\end{figure*}

\begin{figure*}[!htpb]
\begin{adjustbox}{max width=\textwidth}
\begin{tabular}{l@{\ }l@{\ }c@{\ }c@{\ }c@{\ }c}
 \toprule
 & \multicolumn{1}{c}{\huge  Is there an equal number of cubes 
\par} & 
\multicolumn{1}{c}{\huge Are there less matte objects   \par } \\
& \multicolumn{1}{c}{\huge  in front of the large shiny thing  \par} & 
\multicolumn{1}{c}{\huge on the left of the green block \par} \\
& \multicolumn{1}{c}{\huge behind the cyan block\par} & 
\multicolumn{1}{c}{\huge  than metal things?\par}\\
& \multicolumn{1}{c}{\huge and behind the yellow cube? \par} & 
\multicolumn{1}{c}{\huge  \par}
\\\midrule
\rotatebox{90}{\Huge TbD - Before \par} & \multicolumn{1}{c}{\includegraphics[width=0.7\linewidth]{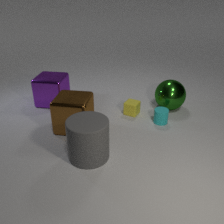}} &
%i mixed up the names %
\includegraphics[width=0.7\linewidth]{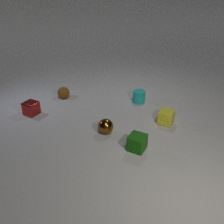}
\\
 & \multicolumn{1}{c}{{\Huge \textcolor{green}{No} \par}} & 
\multicolumn{1}{c}{\Huge \textcolor{green}{Yes} \par}
\\\midrule
\rotatebox{90}{\Huge  TbD - After\par} & \multicolumn{1}{c}{\includegraphics[width=0.7\linewidth]{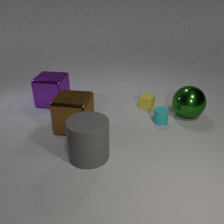}} &
\includegraphics[width=0.7\linewidth]{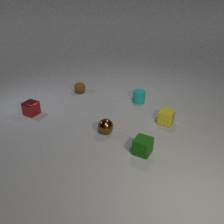}
\\
 & \multicolumn{1}{c}{{\Huge \textcolor{red}{Yes} \par}} & 
\multicolumn{1}{c}{\Huge \textcolor{red}{No} \par}
\\\bottomrule
\end{tabular}
\end{adjustbox}

\caption{Manipulations of CLEVR models. We show results \textit{before} and \textit{after} scene manipulations.}
\label{fig:tbd_qualitative}
\end{figure*}

In Figures~\ref{fig:san_qualitative}-\ref{fig:tbd_qualitative}, we provide more qualitative results.

\clearpage
\subsection{Quantitative Results with p-values}
\label{sec:quantitative_p_value}
We extend the quantitative results of the main paper by also reporting $p$-values of the one-sample hypothesis testing.
We use  the \textit{Accuracy Drop} metric that measures the accuracy drop after the \vqaagent is manipulated, and the \textit{Consistency Drop} that measures how many times the manipulated \vqaagent changes its answer, independently if that is a correct or wrong answer. The higher \textit{Drop Accuracy} or \textit{Consistency Drop}, the more successful manipulations are.
We use T-Test~\citep{helmert1876genauigkeit, luroth1876vergelichung} for each metric all the runs. 
Our null hypothesis is that the population mean of all the games is zero, indicating that a manipulation is unsuccessful, and the results of our \adversarials are extreme cases of a good performance, i.e, $H_0: \mu_{l} = 0, l\in[\text{Drop},\text{Consistency}]$. As an alternative hypothesis, we assume that the population mean is greater than zero, i.e., $H_A: \mu_{l} > 0, l\in[\text{Drop},\text{Consistency}]$. The same setup stands for both the \textit{Consistency} and the \textit{Drop} metric. Tables~\ref{results_state_with_p_value} and~\ref{results_image_with_p_value} show the full table with $p$-values computed. Note that only in two cases, \adversarial does not manipulate the scene convincingly (TbD and Mdetr for the \textit{pixel-input} \adversarial).

\begin{table}[!htbp]
\caption{Results of the game with the \textit{state-input} \adversarial.}
\label{results_state_with_p_value}
\begin{center}
\small
\begin{tabular}{p{1cm}p{1.5cm}p{1.5cm}p{1.5cm}p{1.5cm}p{1.5cm}}
\toprule
\multicolumn{1}{c}{} & \multicolumn{1}{|c}{} & \multicolumn{2}{|c|}{Average}  & \multicolumn{2}{c|}{$p$-value}\\ 
Model    &   \minigame size     &   Consistency Drop \% & Accuracy  Drop \%& Consistency Drop & Accuracy  Drop  \\
 &  &                  &           &                             &                           \\
\midrule
 
SAN  & 10      &           -23.9 &    -14.2 &                                              0.000 &                    0.000 \\
    & 100    &           -10.7 &     -8.3 &                                        0.000 &                    0.000 \\
    & 1000 &                    -3.5 &     -2.5 &                                           0.001 &                    0.000 \\\cmidrule{1-6}
FiLM  & 10      &          -14.8 &    -12.7 &                      0.003 &     0.006 \\
    & 100       &            -7.8 &     -7.3 &                                           0.000 &                    0.000 \\
    & 1000 &              -3.1 &     -2.4 &                                               0.000 &                    0.000 \\\cmidrule{1-6}
    
RN  & 10      &           -20.9 &    -13.6 &                                            0.000 &                    0.000 \\
    & 100      &           -10.5 &     -8.2 &                                              0.000 &                    0.000 \\
    & 1000     &            -3.3 &     -2.8 &                                           0.000 &                    0.000 \\\cmidrule{1-6}

IEP & 10       &           -13.9 &    -13.0 &                                             0.001 &                    0.003 \\
    & 100      &            -6.9 &     -6.8 &                0.000 &                    0.000 \\
    & 1000     &            -2.9 &     -2.8 &                   0.000 &                    0.000 \\\cmidrule{1-6}

TbD & 10       &            -5.3 &     -5.1 &               0.011 &                    0.012 \\
    & 100     &     3.8        &      -2.5 &                  0.000 &                    0.000 \\
    & 1000     &            -1.1 &     -1.1 &                0.000 &                    0.000 \\\cmidrule{1-6}

Mdetr  & 10      &           -6.6 &    -6.0 &                 0.018 &                   0.021 \\
    & 100  &                -4.9 &     -3.6 &              0.000 &                    0.000 \\
    & 1000 &               -1.3 &     -1.2 &                0.000 &                    0.000 \\\cmidrule{1-6}
    
State & 10   &  -8.5 &     -7.7 &            0.020 &                    0.034 \\
Input    & 100  &            -2.3 &     -2.1 &             0.000 &                    0.000 \\
Model    & 1000 &            -1.2 &     -1.1 &             0.000 &                    0.000 \\

\bottomrule
\end{tabular}
\end{center}
\end{table}
\begin{table}[!htpb]
\caption{Results of the game with the \textit{pixel-input} \adversarial.}
\label{results_image_with_p_value}
\begin{center}
\small
\begin{tabular}{p{1.5cm}p{1.5cm}p{1.5cm}p{1.5cm}p{1.5cm}p{1.5cm}}
\toprule
\multicolumn{1}{c}{} & \multicolumn{1}{|c}{} & \multicolumn{2}{|c|}{Average} &  \multicolumn{2}{c|}{$p$-value}\\ 
Model    &   \minigame size  &  Consistency Drop \% & Accuracy  Drop \%& Consistency Drop & Accuracy  Drop  \\
 &  &                  &           &                                   &                           \\
\midrule
SAN  & 10   &           -12.8 &    -11.4 &                                  0.008 &                    0.018 \\
    & 100  &            -6.4 &     -3.9 &                                      0.000 &                    0.001 \\
    & 1000 &            -2.3 &     -1.4 &                                      0.000 &                    0.000 \\\cmidrule{1-6}
    
FiLM  & 10   &            -3.62 &     -2.6 &                                   0.056 &                    0.154 \\
    & 100  &       -1.7 &        -1.4 &                                  0.000 &                    0.000 \\
    & 1000 &            -0.6 &     -0.4 &                                       0.000 &                    0.000 \\\cmidrule{1-6}
    
RN  & 10   &            -9.4 &     -7.2 &                                   0.021 &                    0.065 \\
    & 100  &            -3.2 &     -2.5 &                                     0.000 &                    0.000 \\
    & 1000 &            -2.3 &     -1.7 &                                      0.000 &                    0.000 \\\cmidrule{1-6}

IEP  & 10   &           -2.9 &     -2.5 &                                      0.111 &                    0.171 \\
    & 100  &            -1.5 &     -1.3 &                                         0.001 &                    0.004 \\
    & 1000 &            -0.4 &     -0.3 &                                     0.000 &                    0.000 \\\cmidrule{1-6}
    
TbD  & 10   &  0.0 &     0.0 &                                   1.000 &                   1.000 \\
    & 100  &            -0.6 &     -0.4 &                                    0.017 &                    0.078 \\
    & 1000 &            -0.2 &     -0.1 &                                     0.000 &                    0.000 \\\cmidrule{1-6}
Mdetr  & 10   &            0.0 &     0.0 &                                 1.000 &                    1.000 \\
    & 100  &            -1.2 &     -1.1 &                                    0.002 &                    0.003 \\
    & 1000 &            -0.4 &     -0.3 &                                     0.000 &                    0.000  \\\cmidrule{1-6}
    
State & 10   &  -0.9 &     -0.8 &            0.033 &                    0.038 \\
Input    & 100  &            -0.7 &     -0.7 &               0.031 &                    0.031 \\
Model    & 1000 &            -0.3 &     -0.3 &              0.000 &                    0.000 \\
\bottomrule
\end{tabular}
\end{center}
\end{table}

\clearpage

\section{Appendix B}
\subsection{Model performance on maximal drop minigames.}
In the table below, we present model accuracy on each of the \minigames where the maximal performance drop was observed. The  third and fourth columns (Original Data section) refer to the model accuracy on unmanipulated image-question pairs of the respective \minigame. The fifth and sixth columns (Manipulated Data section) refer to the model accuracy in those \minigame examples after the \adversarial manipulation. 

\begin{table}[!htbp]
\centering
\small
\begin{tabular}{p{1cm} p{2cm} p{1.5cm} p{1.5cm} p{1.5cm} p{1.5cm}}
\toprule
\multicolumn{1}{c}{} & \multicolumn{1}{c|}{} & \multicolumn{2}{|c|}{Original Data} & \multicolumn{2}{|c|}{Manipulated Data}\\ 

\vspace{5pt}Model &\vspace{5pt}  \minigame size   & \vspace{5pt} \minigame Accuracy (State) &\vspace{5pt} \minigame Accuracy (Pixel) & \vspace{5pt} \minigame Accuracy (State) & \vspace{5pt} \minigame Accuracy  (Pixel) \\
 &  &                  &     &                            \\
\midrule
   
SAN  & 10       &      80.00  &  80.00 &  28.84  & 43.26\\  
    & 100       &     74.00  &  75.00 &  47.65 &  66.32\\
    & 1000      &     72.30  &  72.20 &   68.42 & 70.87\\\cmidrule{1-6}

FiLM  & 10      &     100.0  &   100.0 &  48.10 &  86.58\\
    & 100       &      98.00   &   100.0 &   75.61 & 92.44\\
    & 1000      &      96.40   &   96.10 &  90.81 & 94.75\\\cmidrule{1-6}

RN  & 10       &    100.0 &   100.0  &  46.60 &  74.56 \\
    & 100      &     94.00 &   95.00 &   63.28 &  87.42\\
    & 1000     &      93.10 &   93.30 &   90.03 & 91.05\\\cmidrule{1-6}

IEP & 10       &    100.0   &  100.0 &  48.45 &  87.21\\
    & 100      &    98.00   &  96.00 & 74.41 & 93.50\\
    & 1000     &     97.10   &  97.30 & 93.99 & 95.73\\\cmidrule{1-6}

TbD & 10       &      100.0  &   100.0 & 69.37 & 99.10\\
    & 100      &      100.0  &   100.0 &  91.17 &  98.00\\
    & 1000     &       99.40  &   99.70 & 95.53 & 98.80\\\cmidrule{1-6}

Mdetr  & 10     &     100.0   &  100.0 & 59.82 & 99.70\\
    & 100       &   100.0   &  100.0 &  86.43 & 97.80\\
    & 1000      &    99.50   &  100.0 & 94.51 & 99.10\\\cmidrule{1-6}
    
State & 10       &     100.0    &   100.0 & 77.44 &  91.96\\
Input  & 100     &     97.00    &   99.00 & 92.63 &  95.05\\
Transf. & 1000   &      97.20   &   96.40 & 95.15 &  95.83\\

\bottomrule
\end{tabular}
\label{table:avg_per}
\end{table}

\subsection{Generating `Visually Fair' Scenes.}
\label{sec:vis_fair}
We define a scene as \textit{visually fair} if there are no occlusions and all objects are within the field-of-view. The ground truth answer of a posed question on a \textit{visually fair} scene remains the same before and after object manipulation.

For the generation of new images/scenes we use the open-source Blender Graphics Engine \footnote{Blender - a 3D modeling and rendering package. \url{https://www.blender.org}} (v2.79b), and the original 3D models of the CLEVR dataset.
Scenes suggested by our \adversarial, are encoded in the form of arguments for Blender, which renders the new images according to those directives. 
In order to guarantee visual fairness, and question validity, we employ two modules (scene-constraint enforcer / question relevance enforcer), presented  briefly - in Section 4 in the Consistency and In-Distribution paragraphs. 

The scene-constraint enforcer is responsible for ensuring the creation of a scene/image that
respects all constraints and statistical properties of the original dataset. This piece of code is borrowed from the original CLEVR codebase\footnote{\url{ https://github.com/facebookresearch/clevr-dataset-gen/blob/main/image_generation/render_images.py}}  and as in the original CLEVR paper the following constraints are checked  (Lines 66-83 in the python code):

\begin{enumerate}
    \item \textbf{Minimum / Maximum Number of Objects:} 3 / 10.
    \item \textbf{Minimum allowed distance between object centers:} 0.25
    \item \textbf{Margin along with cardinal directions:} 0.4 (This ensures lack of ambiguity by keeping objects at least 0.4 points of distance apart. This distance is a Blender-specific measure that spans from -3 to +3.)
    \item \textbf{Min pixels per object:} 200 ( This ensures that an object will not be occluded, by requiring at least 200 pixels of it being visible.)
\end{enumerate}

The question-relevance enforcer checks if the newly generated scene is indeed answerable and has the same “ground truth” answer as the old scene. By not enforcing this, every manipulation will be falsely “successful” as the question (which is kept the same before and after the attack) will have a different answer in the newly generated scene. 

This is also resolved by using the same question engine module provided by the CLEVR authors in their codebase\footnote{\url{https://github.com/facebookresearch/clevr-dataset-gen/blob/main/question_generation/question_engine.py}}.
The module receives a question and a scene and calculates the ground truth answer based on a set of functional programs that are apriori known and used to generate the questions. (Supplementary Material A in the CLEVR paper \citep{Johnson_2017_CVPR})

We use the following pipeline.
Our \adversarial initially receives an input image-question pair.
Then, it suggests a new scene, that passes through the scene-constraint enforcer. If the new scene respects all boundaries and constraints it is then evaluated alongside the question by the question-relevance enforcer.
If the new ground truth answer to the suggested scene/image is the same as the original answer, the image-question pair is marked as in-distribution and valid. Otherwise, it is removed from the training pipeline.

\newpage

\subsection{Comparison with randomly sampled equivalent scenes.}
Arguably, out of the semantically equivalent space of images given a question and an accompanying answer, the \adversarial  effectively discovers cases for which the model under the test can be fooled. This, however, does not indicate how rare those examples are, neither the degree of robustness of the model against other examples in this space. Let us limit ourselves on a specific image-question pair. Then, a tractable estimate of how rare "fooling examples" are would be the number of manipulated scenes ($F$)  that are able to fool the model, divided by all the possible permutations of available objects ($N$) in our discrete 7x7 grid. $Rarity = \frac{F}{49^N}, N \in [3,10]$. A scene of only 3 objects and 1 discovered successful manipulation would make it 0.00085\% possible to find it by chance and would indicate 99.9915\% robustness against that sphere of equivalent image-question pairs.

Of course not all of those configurations are valid  due to scene constraint violations, as well as objects can theoretically move freely on the scene. As a way to approach this question, we compare our \adversarial  against a random scene generator (RSG). The random scene generator is able to manipulate any object freely, under the constraint that the resulting scene is valid under the scene constraint enforcer as well as the question relevance enforcer. Manipulations of lighting or camera angles, as well as addition or removal of objects is not permitted. In order to reduce the huge rendering times needed for each random search, we allow the RSG to operate on a 10 length \minigame where scene manipulations that cause accuracy drop have been discovered by the \adversarial. For each image-question pair the RSG has a budget of 5000 queries, and if not successful it proceeds to the next \minigame entry. 

\begin{table}[!htbp]
\begin{tabular}{|l|l|l|l|}
\cline{1-4}
Model                       & CLEVR Accuracy & Adv.Player Accuracy (Drop \%) & RSG Accuracy (Drop \%) \\ \hline
\multicolumn{1}{|l|}{SAN}   & 72.1           & 61.86 (-14.2\%)               & 71.59 (-0.7\%)         \\ \hline
\multicolumn{1}{|l|}{FiLM}  & 96.2           & 83.98 (-12.7\%)               & 95.81 (-0.4\%)         \\ \hline
\multicolumn{1}{|l|}{RN}    & 93.2           & 80.52 (-13.6\%)               & 92.73 (-0.5\%)         \\ \hline
\multicolumn{1}{|l|}{IEP}   & 96.9           & 84.30 (-13.0\%)               & 96.70 (-0.2\%)         \\ \hline
\multicolumn{1}{|l|}{TbD}   & 99.1           & 94.04 (-5.1\%)                & 99.10 (-0.0\%)         \\ \hline
\multicolumn{1}{|l|}{MDetr} & 99.7           & 93.71 (-6.0\%)                & 99.70 (-0.0\%)         \\ \hline
\multicolumn{1}{|l|}{ST}    & 96.8           & 89.34 (-7.7\%)                & 96.50 (-0.3\%)         \\ \hline
\end{tabular}
\label{table:rand_per}
\end{table}

As it can be seen, randomly question-uninformed suggested object placements do not seem to be effective into fooling the models under test. This further supports the argument that models under test theoretically and statistically seem robust, but nevertheless  have vulnerabilities that can be effectively exploited by our \adversarial.

\newpage

\subsection{Visualizations.}
\label{sec:failure}
Here, we try to isolate reasoning gaps from visual causes after the scene manipulations; with the following setup. Each \vqaagent is  introduced with two image-question pairs. The first consists of the original image and the second of a scene created by our \adversarial that successfully fools the \vqaagent. We then inspect the visualization of the \vqaagent's inner-workings in both pairs. We are especially interested in patterns that suggest the root of failure.
As each \vqaagent operates under different assumptions and architectural designs, we choose to visualize :
\begin{enumerate}
    \item The \textit{attention maps} of Stack-Attention Networks.
    \item The \textit{gradient-weighted activations of the pre-ultimate and ultimate convolutional blocks} of FiLM.
    \item The \textit{programs} that are synthesized alongside their \textit{soft-attention maps} of each module block in TbD network.
    \item The \textit{objects} that were detected during the computation steps of a MDetr model.
\end{enumerate}

\begin{figure*}[!htpb]
\begin{adjustbox}{max width=0.95\textwidth}
\begin{tabular}{cccll}
\toprule

& \multicolumn{3}{c}{Is the number of small cubes that are right of the blue rubber block greater than the number of small balls that } \\
& \multicolumn{3}{c}{are in front of the cyan object? \textcolor{green}{Yes} $\longrightarrow$ \textcolor{red}{No}}
\\
\rotatebox{90}{Original Scene} & \includegraphics[width=0.33\linewidth]{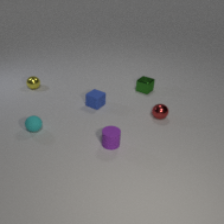} &
\includegraphics[width=0.33\linewidth]{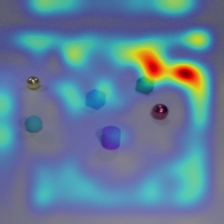} &
\includegraphics[width=0.33\linewidth]{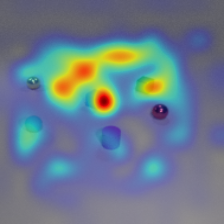} &
\\
 & \multicolumn{1}{c}{} & 
\multicolumn{1}{c}{} & 
\multicolumn{1}{c}{} &
\multicolumn{1}{c}{}
\\
\rotatebox{90}{Manipulated Scene} & \includegraphics[width=0.33\linewidth]{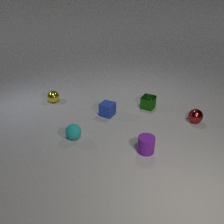} &
\includegraphics[width=0.33\linewidth]{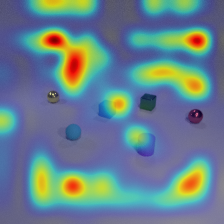} &
\includegraphics[width=0.33\linewidth]{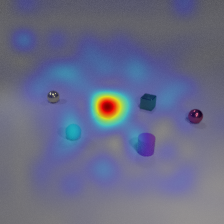} &
\\
 & \multicolumn{1}{c}{} & 
\multicolumn{1}{c}{} & 
\multicolumn{1}{c}{} &
\multicolumn{1}{c}{}
\\\midrule

& \multicolumn{3}{c}{What shape is the thing that is in front of the small } \\
& \multicolumn{3}{c}{yellow cylinder and to the right of the rubber thing? \textcolor{green}{Cylinder} $\longrightarrow$ \textcolor{red}{Sphere}}

\\
\rotatebox{90}{Original Scene} & \includegraphics[width=0.33\linewidth]{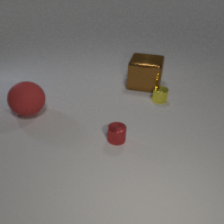} &
\includegraphics[width=0.33\linewidth]{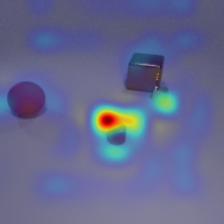} &
\includegraphics[width=0.33\linewidth]{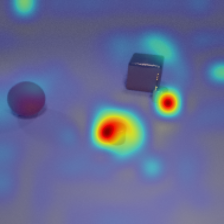} &
\\
 & \multicolumn{1}{c}{} & 
\multicolumn{1}{c}{} & 
\multicolumn{1}{c}{} &
\multicolumn{1}{c}{}
\\
\rotatebox{90}{Manipulated Scene} & \includegraphics[width=0.33\linewidth]{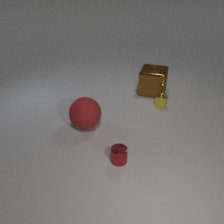} &
\includegraphics[width=0.33\linewidth]{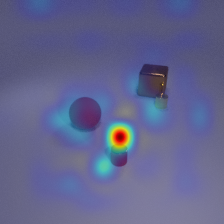} &
\includegraphics[width=0.33\linewidth]{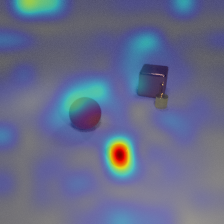} &
\\
 & \multicolumn{1}{c}{} & 
\multicolumn{1}{c}{} & 
\multicolumn{1}{c}{} &
\multicolumn{1}{c}{}
\\\bottomrule
\end{tabular}
\end{adjustbox}

\caption{Visualization of attention maps in SAN model on original and manipulated scenes.}
\label{fig:san_visualisation}
\end{figure*}

In \Figref{fig:san_visualisation}, we visualize the pre-ultimate (column 2) and ultimate (column 3) attention maps of the Stack-Attention Network. We observe that the model focuses on a single final object in manipulated examples. In the original ones, the relational jumps required for the answer are still visible on the attention map. The final attention map that is presented (column 3) is passed to a fully-connected classifier in order for the answer to be created. We observe limited focus to all the necessary objects / locations that creates a lack of necessary features for the correct answer derivation.

\begin{figure*}[!htpb]
\begin{adjustbox}{max width=0.95\textwidth}
\begin{tabular}{l@{\ }c@{\ }c@{\ }c@{\ }c}
 \toprule
& \multicolumn{3}{c}{How many other objects are the same shape as } \\
& \multicolumn{3}{c}{the big gray matte object? \textcolor{green}{1} $\longrightarrow$ \textcolor{red}{0}}
\\
\rotatebox{90}{Original Scene} & \multicolumn{1}{c}{\includegraphics[width=0.33\linewidth]{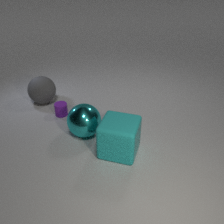}} &
\includegraphics[width=0.33\linewidth]{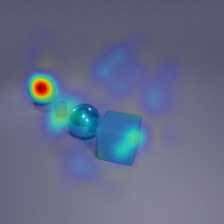} &
\includegraphics[width=0.33\linewidth]{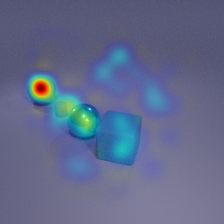} &
\\
 & \multicolumn{1}{c}{} & 
\multicolumn{1}{c}{} & 
\multicolumn{1}{c}{} &
\multicolumn{1}{c}{}
\\
\rotatebox{90}{Manipulated Scene} & \multicolumn{1}{c}{\includegraphics[width=0.33\linewidth]{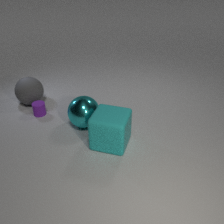}} &
\includegraphics[width=0.33\linewidth]{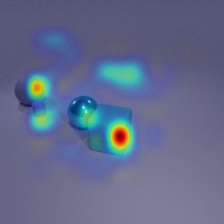} &
\includegraphics[width=0.33\linewidth]{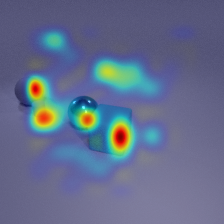} &
\\
 & \multicolumn{1}{c}{} & 
\multicolumn{1}{c}{} & 
\multicolumn{1}{c}{} &
\multicolumn{1}{c}{}
\\\midrule
& \multicolumn{3}{c}{How many other things are the same shape as } \\
& \multicolumn{3}{c}{the big metallic thing? \textcolor{green}{1} $\longrightarrow$ \textcolor{red}{2}}
\\
\rotatebox{90}{Original Scene} & \multicolumn{1}{c}{\includegraphics[width=0.33\linewidth]{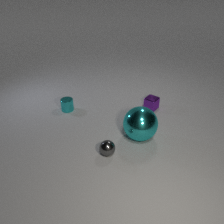}} &
\includegraphics[width=0.33\linewidth]{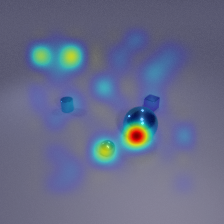} &
\includegraphics[width=0.33\linewidth]{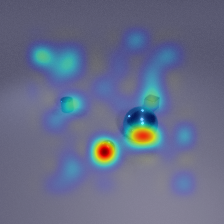} &
\\
 & \multicolumn{1}{c}{} & 
\multicolumn{1}{c}{} & 
\multicolumn{1}{c}{} &
\multicolumn{1}{c}{}
\\
\rotatebox{90}{Manipulated Scene} & \multicolumn{1}{c}{\includegraphics[width=0.33\linewidth]{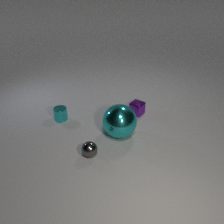}} &
\includegraphics[width=0.33\linewidth]{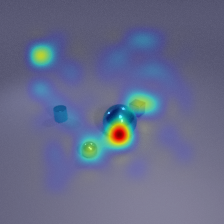} &
\includegraphics[width=0.33\linewidth]{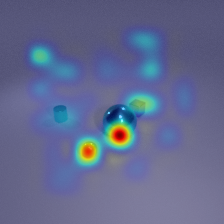} &
\\
 & \multicolumn{1}{c}{} & 
\multicolumn{1}{c}{} & 
\multicolumn{1}{c}{} &
\multicolumn{1}{c}{}
\\\bottomrule
\end{tabular}
\end{adjustbox}

\caption{Gradient weighted activation visualization of FiLM model's pre-ultimate and ultimate blocks on original and manipulated scenes}
\label{fig:grad_vis}
\end{figure*}

In the case of FiLM model in \Figref{fig:grad_vis}, we observe a similar scenario. In the first example, (rows 1 and 2) the gradient-weighted activations of the model seem to be focusing on multiple irrelevant objects regarding the final answer, while in the second example (rows 3 and 4) the model seems to include the purple cube mistakenly as one of the possible spheres of the image. In both cases above it is difficult to exactly pin down the root of (reasoning or visual) failures.

\begin{figure*}[!htpb]
\begin{adjustbox}{max width=0.95\textwidth}
\begin{tabular}{l@{\ }c@{\ }c@{\ }c}
 \toprule
& \multicolumn{3}{c}{The red thing that is the same material} \\
& \multicolumn{3}{c}{as the brown cube is what size? \textcolor{green}{Large} $\longrightarrow$ \textcolor{red}{Small}}
\\
\rotatebox{90}{Original Scene} & \multicolumn{1}{c}{\includegraphics[width=0.5\linewidth]{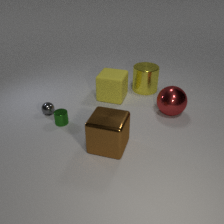}} &
\includegraphics[width=0.5\linewidth]{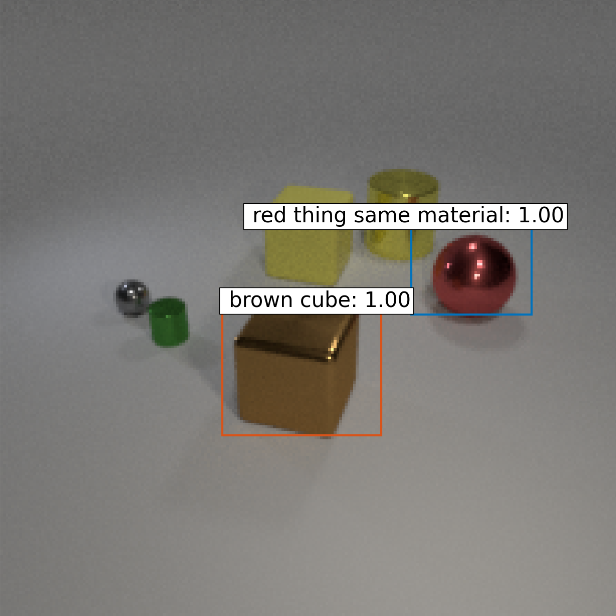} &
\\
 & \multicolumn{1}{c}{} & 
\multicolumn{1}{c}{} 
\\
\rotatebox{90}{Manipulated Scene} & \multicolumn{1}{c}{\includegraphics[width=0.49\linewidth]{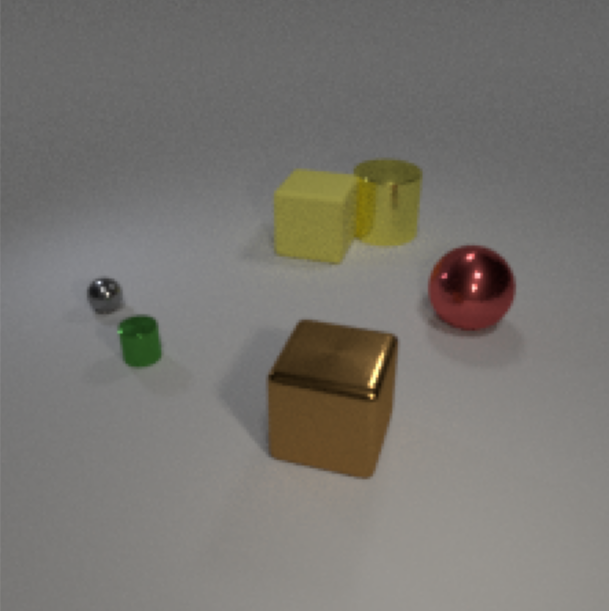}} &
\includegraphics[width=0.5\linewidth]{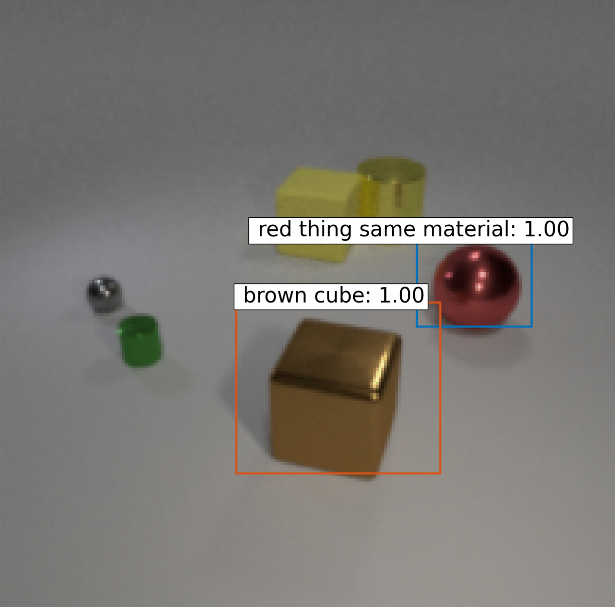} &
\\
 & \multicolumn{1}{c}{} & 
\multicolumn{1}{c}{} 
\\\bottomrule
\end{tabular}
\end{adjustbox}

\caption{Visualization of MDetr model's detection stages on original and manipulated scenes.}
\label{fig:object_vis}
\end{figure*}

\begin{figure*}[!htpb]
\begin{adjustbox}{max width=0.95\textwidth}
\begin{tabular}{l@{\ }c@{\ }c@{\ }c}
 \toprule
& \multicolumn{3}{c}{Do the object behind the big matte sphere and the} \\
& \multicolumn{3}{c}{thing on the right side of the red thing} \\
& \multicolumn{3}{c}{have the same material?
\textcolor{green}{Yes} $\longrightarrow$ \textcolor{red}{No}}
\\
\rotatebox{90}{Original Scene} & \multicolumn{1}{c}{\includegraphics[width=0.5\linewidth]{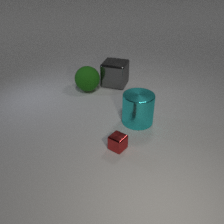}} &
\includegraphics[width=0.5\linewidth]{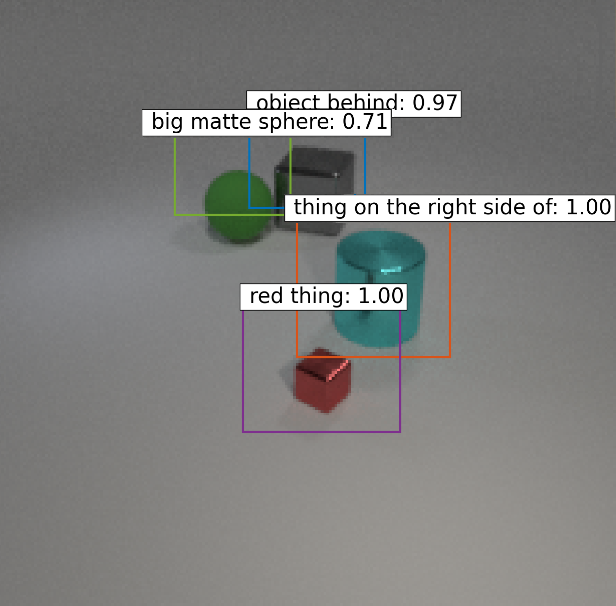} &
\\
 & \multicolumn{1}{c}{} & 
\multicolumn{1}{c}{} 
\\
\rotatebox{90}{Manipulated Scene} & \multicolumn{1}{c}{\includegraphics[width=0.5\linewidth]{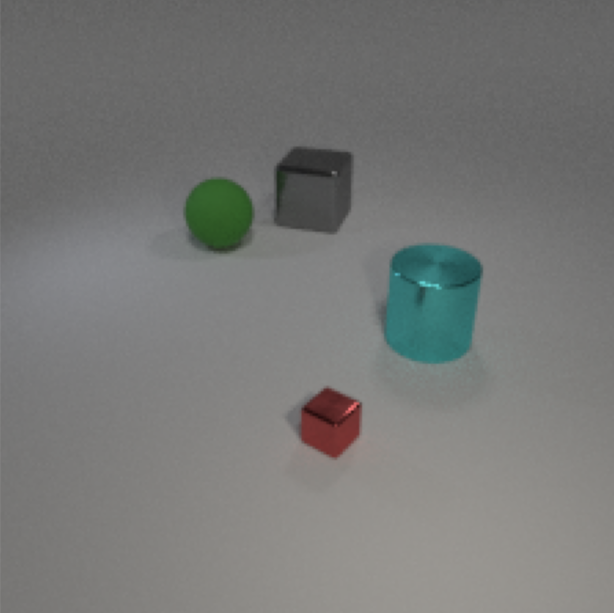}} &
\includegraphics[width=0.5\linewidth]{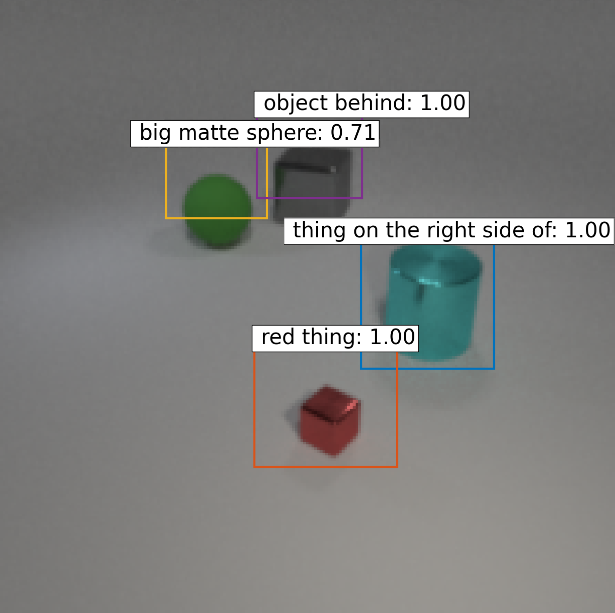} &
\\
 & \multicolumn{1}{c}{} & 
\multicolumn{1}{c}{} 
\\\bottomrule
\end{tabular}
\end{adjustbox}

\caption{Visualization of MDetr model's detection stages on original and manipulated scenes.}
\label{fig:object_vis2}
\end{figure*}

MDetr model can be seen in \Figref{fig:object_vis} and \Figref{fig:object_vis2}. Here, a specific pattern could not be identified. All objects were correctly classified. It suggests that the cause of failures stem rather from the reasoning steps.

\begin{figure*}[!htpb]
\begin{adjustbox}{max width=1.5\textwidth}
\begin{tabular}{l@{\ }c@{\ }c@{\ }c@{\ }c@{\ }c}
 \toprule
& \multicolumn{5}{c}{Is there a big brown object of the same shape as the green thing? \textcolor{green}{Yes} $\longrightarrow$ \textcolor{red}{No}}
\\
\rotatebox{90}{Original Scene} & \multicolumn{1}{c}{\includegraphics[width=\linewidth, height=0.2\linewidth]{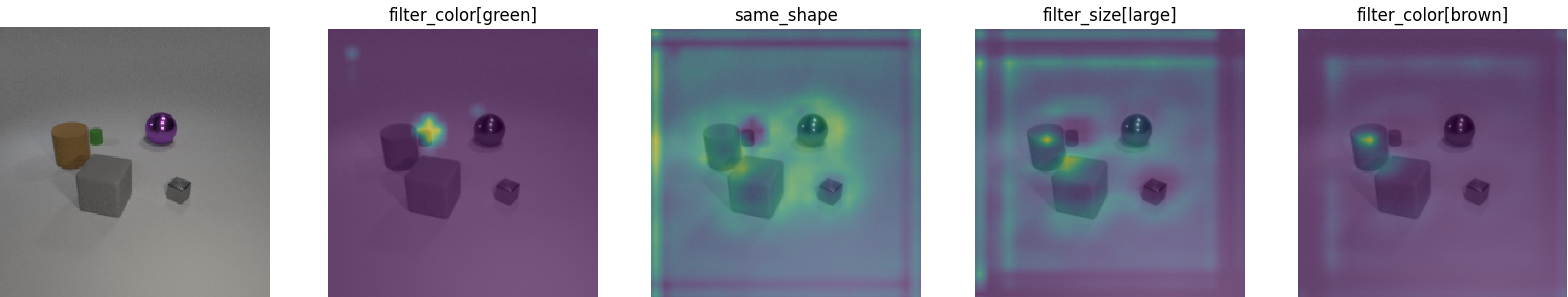}} 
\\
\rotatebox{90}{Manipulated Scene} & \multicolumn{1}{c}{\includegraphics[width=\linewidth,height=0.2\linewidth]{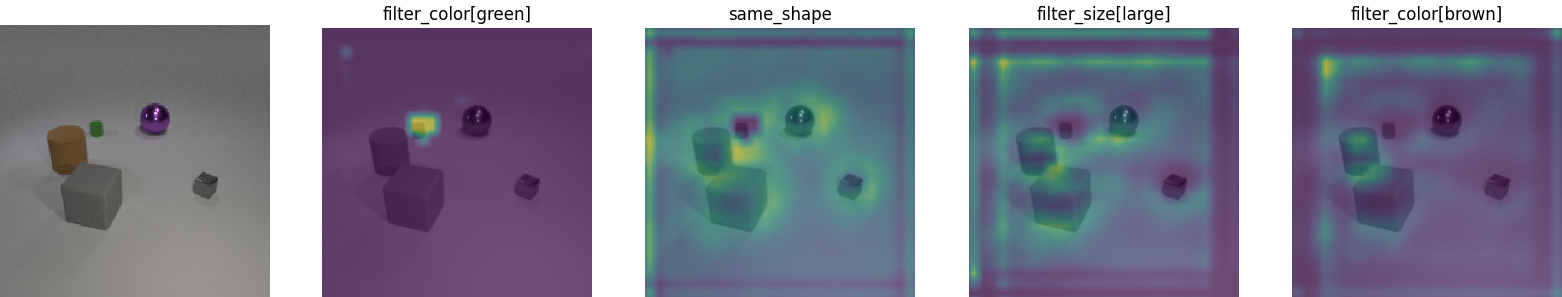}} 
\\\midrule
& \multicolumn{5}{c}{Are there any other things that have the same material as the big yellow ball? \textcolor{green}{Yes} $\longrightarrow$ \textcolor{red}{No}} 
\\
\rotatebox{90}{Original Scene} & \multicolumn{1}{c}{\includegraphics[width=\linewidth,height=0.2\linewidth]{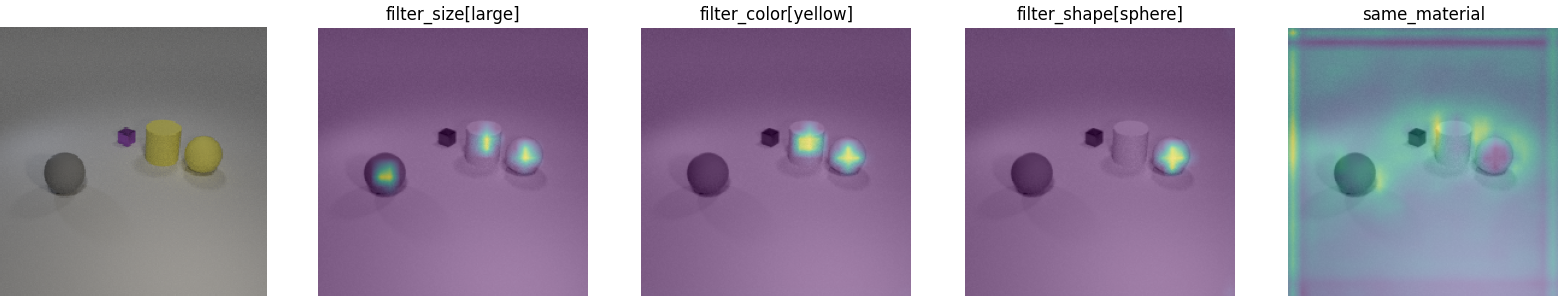}}
\\
\rotatebox{90}{Manipulated Scene} & \multicolumn{1}{c}{\includegraphics[width=\linewidth,height=0.2\linewidth]{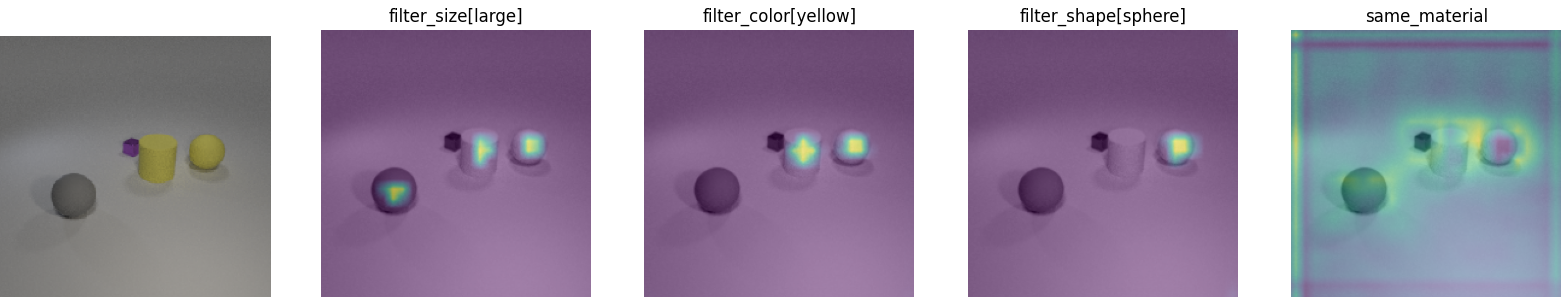}} 
\\\bottomrule
\end{tabular}
\end{adjustbox}

\caption{Visualization of TbD model's program generation and attention map visualization on original and manipulated scenes}
\label{fig:progam_vis}
\end{figure*}

Tbd is presented in \Figref{fig:progam_vis}. In both the original and manipulated scenes the programs that the model generates are exactly the same. This is to no surprise since the program generator module is using exclusively the question as input.
Nevertheless, a similar pattern of scattered model focus is presented here as well.
The attention maps in both examples are the same between the original and manipulated scenes, in all but the final reasoning step.
In the first example (rows 1 and 2) the attention is leaked towards the gray cube that has nothing to do with the correct answer, while in the second (rows 3 and 4) the attention is spread among many matte objects but peaks near the big yellow ball. This is wrong since the word "other" that exists in the question should drive the model focus exclusively to other objects apart from the starting point of the reasoning chain.

\clearpage

\subsection{Human-in-the-loop Experiment}
\label{sec:human}
\begin{figure*}[!htpb]
\centering
\includegraphics[width=0.8\textwidth]{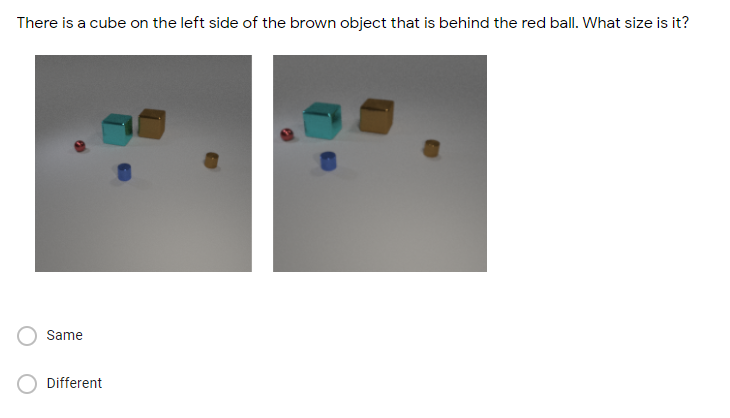}
\caption{Questionnaire example: The observer is presented with two images and a question. Then they have to pick if their answer would remain the same if the question referred to each image respectively.}
\label{fig:survey}
\end{figure*}

\begin{figure*}[!htpb]
\centering
\includegraphics[width=0.5\textwidth]{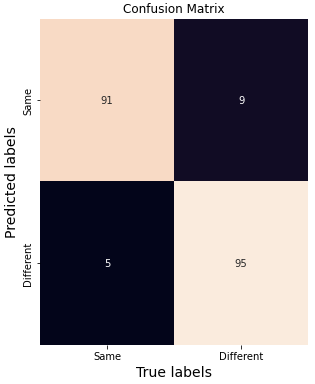}
\caption{Confusion matrix of human experiment: Participants identified 91\% of cases where manipulations were not causing a change in the ground-truth answer. All of these managed to fool \vqaagents and made up half of the survey. When faced with manipulations that changed the ground-truth answer (the other half of the survey), they identified 95\% of them.}
\label{fig:confmat}
\end{figure*}

Can our \adversarial `fool' a human-agent? The perceptual abilities of human observers have been previously measured (92.6\%) on this dataset, however, human robustness is a  different question. We, hence, decided to assess human robustness by a simple experiment. In our setting, a human observer is presented with a set of two images and a question. Then the observer is required to answer to each image question pair, (image before - question / image after - question), and report whether the answer has changed or remained the same. Participants were first made familiar with the concepts of the CLEVR dataset (colors / shapes / materials / sizes) and then were left unsupervised to answer a set of 20 such examples.
Each contained one CLEVR question and two CLEVR scenes, before and after the \adversarial manipulations. Half of those scenes successfully fooled models and half of those did not. An example of what a survey question looks like is given in \Figref{fig:survey}, together with the confusion matrix \Figref{fig:confmat} (results of our survey).

As we can see in \Figref{fig:confmat},  human observers exhibit 91\% precision in identifying cases where scene manipulations are not causing changes to the questions answer, and 95\% precision in identifying cases where the manipulations do cause a change. While this small scale experiment may not reflect a universal truth about human robustness in VQA scenarios, it is a suggestion that human observation is still reliable after our agent's manipulations.
\newpage
\subsection{Object Detection Experiment}
\label{sec:obj_detect}
\begin{figure*}[!htpb]
\begin{adjustbox}{max width=1\textwidth, max height=135pt}
\begin{tabular}{l@{\ }c@{\ }c@{\ }c}
 \toprule
\\
\rotatebox{90}{Original Scene} & \multicolumn{1}{c}{\includegraphics[width=0.5\linewidth]{images/visfail/CLEVR_val_003000.png}} & \includegraphics[width=0.5\linewidth]{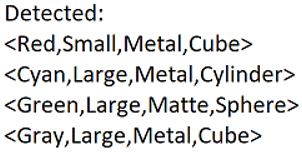}

\\
 & \multicolumn{1}{c}{} & 
\multicolumn{1}{c}{} 
\\
\rotatebox{90}{Manipulated Scene} & \multicolumn{1}{c}{\includegraphics[width=0.5\linewidth]{images/visfail/mdert_val_003000.png}} & \includegraphics[width=0.5\linewidth]{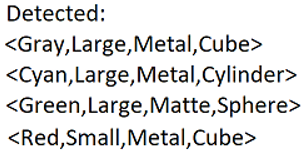}
\\
 & \multicolumn{1}{c}{} & 
\multicolumn{1}{c}{} 
\\\bottomrule
\end{tabular}
\end{adjustbox}

\caption{Visualization of Slot Attention model's detection results}
\label{fig:object_sa}
\end{figure*}

\begin{figure*}[!htpb]
\begin{adjustbox}{max width=1\textwidth, max height=135pt}
\begin{tabular}{l@{\ }c@{\ }c@{\ }c}
 \toprule
\\
\rotatebox{90}{Original Scene} & \multicolumn{1}{c}{\includegraphics[width=0.5\linewidth]{images/visfail/CLEVR_val_000541.png}} &\includegraphics[width=0.5\linewidth]{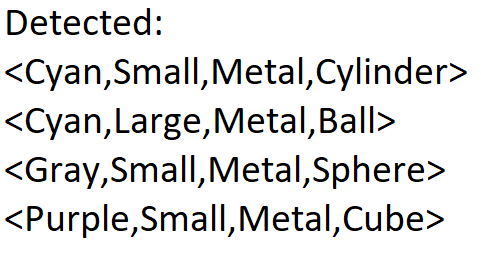}

\\
 & \multicolumn{1}{c}{} & 
\multicolumn{1}{c}{} 
\\
\rotatebox{90}{Manipulated Scene} & \multicolumn{1}{c}{\includegraphics[width=0.5\linewidth]{images/visfail/film_val_000541.png}} & \includegraphics[width=0.5\linewidth]{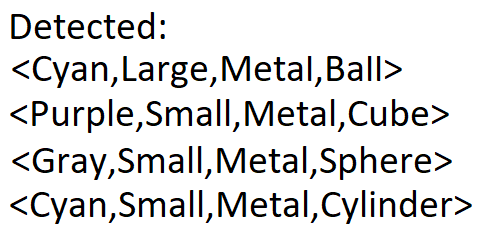}
\\
 & \multicolumn{1}{c}{} & 
\multicolumn{1}{c}{} 
\\\bottomrule
\end{tabular}
\end{adjustbox}

\caption{Visualization of Slot Attention model's detection results}
\label{fig:object_sa2}
\end{figure*}

As an extra test to confirm the visual fairness (see \secref{sec:vis_fair}) of \adversarial manipulations we run the following experiment.
We employed an unsupervised object discovery model, called Slot Attention \citep{slot}, pre-trained on CLEVR, together with a classifier trained to identify object attributes. 
% It is composed of an unsupervised module of object discovery, and a supervised classifier that identifies object attributes. 
% This model was used on image pairs made up from the original and manipulated versions of scenes. We use pairs in which the manipulated version fooled a \vqaagent. 
Next, we used that model to identify objects and attributes on original and manipulated scenes.
% For each image in a pair, we keep track of all the detected objects and their identified attributes. 
Ultimately, we could not find any pairs in which the number of objects or their attributes were different before and after the manipulations. The results for two pairs can be seen in \Figref{fig:object_sa} and \Figref{fig:object_sa2}.
We thus conclude that our \adversarial manipulations create `visually fair' scenes, without producing any corner cases of perceptual system.

